\documentclass[12pt]{iopart}

\usepackage{amsfonts, amsbsy, amssymb, amsthm, color, url}

\usepackage[mathscr]{eucal}
\usepackage{setstack}

\usepackage{algorithm}
\usepackage{algpseudocode}
\usepackage{xcolor}

\usepackage{multirow, adjustbox}

\usepackage{graphicx}
\DeclareGraphicsExtensions{.pdf,.jpeg,.jpg,.png,.eps}

\begin{document}

\title[Hierarchical Decomposed Dual-domain DL for Sparse-View CT Reconstruction]{Hierarchical Decomposed Dual-domain\\ Deep Learning for Sparse-View CT Reconstruction}

\author{Yoseob Han \textsuperscript{\rm 1, 2}}
\address{\textsuperscript{\rm 1} Department of Electronic Engineering, Soongsil University, South Korea\\
        \textsuperscript{\rm 2} Department of Intelligent Semiconductors, Soongsil University, South Korea\\}
\ead{yoseob.han@ssu.ac.kr}
\vspace{10pt}
\begin{indented}
\item[]December 2023
\end{indented}

\begin{abstract}
Objective: X-ray computed tomography employing sparse projection views has emerged as a contemporary technique to mitigate radiation dose. However, due to the inadequate number of projection views, an analytic reconstruction method utilizing filtered backprojection results in severe streaking artifacts. Recently, deep learning strategies employing image-domain networks have demonstrated remarkable performance in eliminating the streaking artifact caused by analytic reconstruction methods with sparse projection views. Nevertheless, it is difficult to clarify the theoretical justification for applying deep learning to sparse view CT reconstruction, and it has been understood as restoration by removing image artifacts, not reconstruction. 

Approach: By leveraging the theory of deep convolutional framelets and the hierarchical decomposition of measurement, this research reveals the constraints of conventional image- and projection-domain deep learning methodologies, subsequently, the research proposes a novel dual-domain deep learning framework utilizing hierarchical decomposed measurements. Specifically, the research elucidates how the performance of the projection-domain network can be enhanced through a low- rank property of deep convolutional framelets and a bowtie support of hierarchical decomposed measurement in the Fourier domain. 

Main Results: This study demonstrated performance improvement of the proposed framework based on the low-rank property, resulting in superior reconstruction performance compared to conventional analytic and deep learning methods. 

Significance: By providing a theoretically justified deep learning approach for sparse-view CT reconstruction, this study not only offers a superior alternative to existing methods but also opens new avenues for research in medical imaging. It highlights the potential of dual-domain deep learning frameworks to achieve high-quality reconstructions with lower radiation doses, thereby advancing the field towards safer and more efficient diagnostic techniques. The code is available at https://github.com/hanyoseob/HDD-DL-for-SVCT.
\end{abstract}

\clearpage    

\section{Introduction}
X-ray computed tomography (CT) has gained widespread acceptance for its ability to produce high-quality and high-resolution images. However, one critical concern associated with the use of X-ray CT lies in its potential to increase the risk of cancer due to the radiation exposure it entails \cite{shah2008alara}. In response to this concern, many researches have been dedicated to developing strategies for reducing radiation exposure \cite{nuyts2013modelling}. These studies were integrated around three main approaches: 
(1) \textit{\textbf{low-dose CT}}, focusing on photon counts of X-ray source
\cite{leuschner2021lodopab, xia2021magic, chun2019bcd, shan2019competitive, yin2019domain};
(2) \textit{\textbf{interior tomography}}, emphasizing region-of-interest (ROI) \cite{han2022end, han2019one}; and 
(3) \textit{\textbf{sparse-view CT}}, involving projection views
\cite{wu2021drone, lee2020sparse, hu2020hybrid, han2018framing, xie2018artifact}. 
Specifically, conventional multi-detector CT (MDCT), which requires rapid and continuous measurement acquisition, is limited in its ability to use the sparse-view CT. However, the sparse-view CT is intriguing for new applications, including spectral CT using alternating kVp switching \cite{kim2014sparse}, dynamic beam blocker \cite{lee2016moving}, etc. 
Additionally, when applied to C-arm CT or dental CT, the scanning duration is primarily constrained by the relatively slower speed of the flat-panel detector, as opposed to the mechanical speed of the gantry. Therefore, the sparse-view CT offers a promising method for reducing scanning time in these contexts \cite{bian2010evaluation, pan2009commercial}.

However, incomplete projection views collected from sparse-view CT can lead to severe streaking artifacts when applying analytic methods such as filtered backprojection (FBP). To address this issue, researchers have explored the use of compressed sensing (CS) techniques \cite{donoho2006compressed} that minimizes total variation (TV) or other forms of sparsity-inducing penalties under data fidelity term 
\cite{lu2011few, ramani2011splitting, bian2010evaluation, pan2009commercial, sidky2008image}. 
Nevertheless, these methods require significant computational burden due to the need for repeated projection and backprojection operations in iterative update steps.

Over the past few years, deep learning (DL) has emerged as a high-performance algorithm in the field of CT image reconstruction.
These DL-based algorithms have demonstrated superior performance compared to traditional model-based iterative reconstruction (MBIR) methods 
\cite{yu2009compressed, ramani2011splitting}, excelling in both image quality and reconstruction time.

\begin{figure}[!t]
    \centering
    \includegraphics[width=0.7\textwidth]{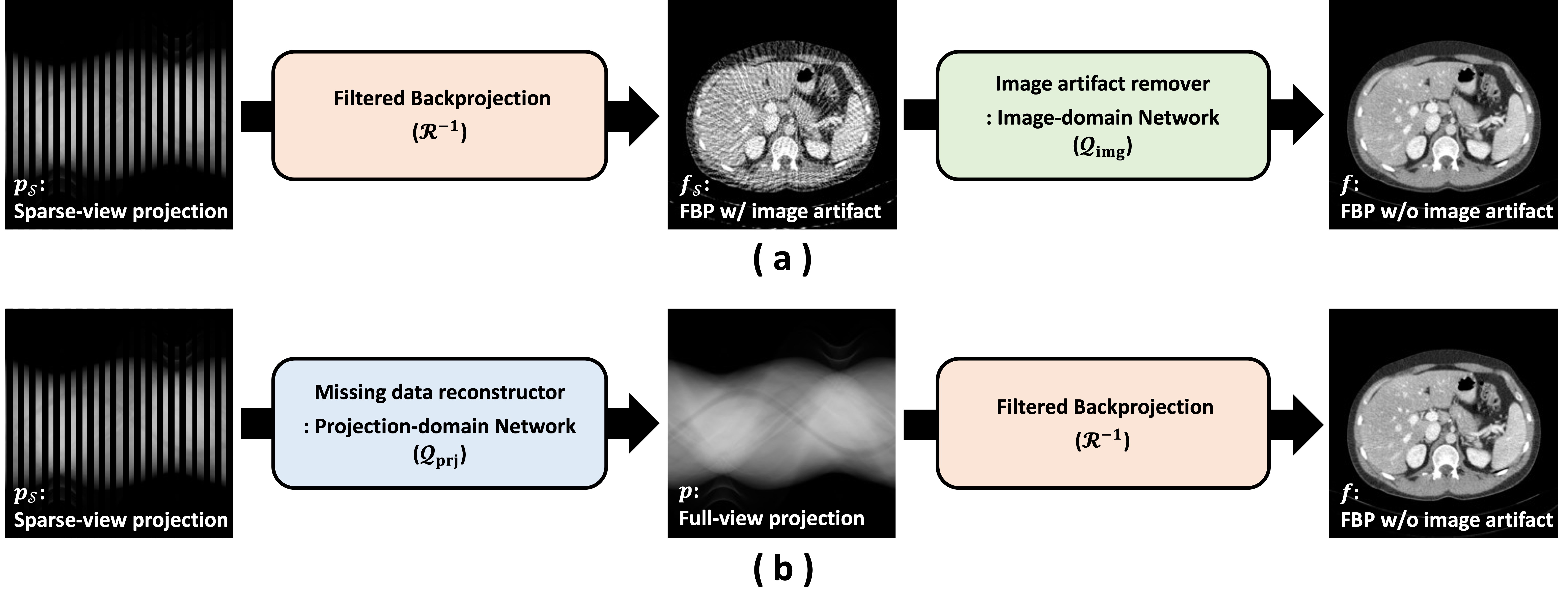}
    \caption{Types of deep learning approaches for sparse-view CT. (a) Image artifact remover using image-domain DL and (b) missing data reconstructor using projection-domain DL.}
    \label{fig:sparse_view}
\end{figure}

\begin{figure}[!t]
    \centering
    \includegraphics[width=1.0\textwidth]{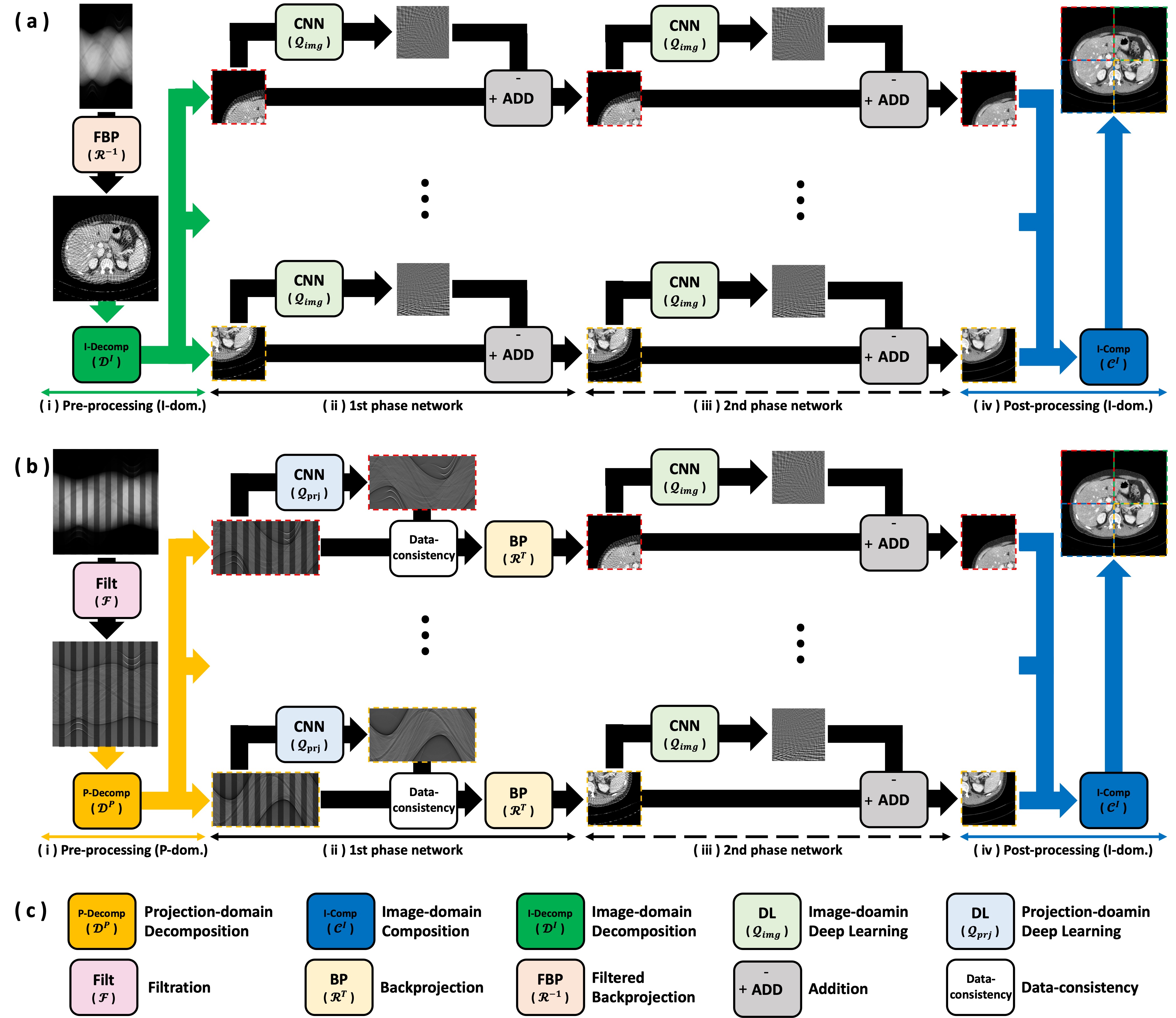}
    \caption{(a) Image-domain DL framework consisting of two image-domains networks and (b) proposed DL framework consisting projection-domain network and image-domain network. (c) Function modules used in (a) and (b). The network has four parts: (i) pre-processing, (ii) 1st phase network, (iii) 2nd phase network, and (iv) post-processing.}
    \label{fig:methods}
\end{figure}

As illustrated in Figure \ref{fig:sparse_view}(a), a sparse-view CT image suffers from streaking artifacts presented as global artifacts. Conventional image-domain DLs 
\cite{han2018framing, lee2018deep, chen2017low, jin2017deep} (see Figure \ref{fig:methods}(a)(ii)) 
work to mitigate streaking artifacts within the image domain. As depicted in Figure \ref{fig:sparse_view}(a), these approaches primarily perform the function as \textit{\textbf{image artifact removers}}, but the underlying cause of artifacts mainly originates from incomplete measurements within the projection domain, such as a limited numbers of views. In addition, the image-domain DL requires the use of entire CT images rather than patch images to effectively capture the global features of the streaking artifact. 
In an effort to directly address the issue of incomplete measurement, projection-domain DLs 
\cite{lee2018deep, dong2019sinogram, wu2022deep, wu2021drone, hu2020hybrid} 
(see Figure \ref{fig:methods}(b)(ii)) have been introduced to reconstruct undersampled projection data. These approaches work as \textit{\textbf{missing data reconstructor}}, as illustrated in Figure \ref{fig:sparse_view}(b). To train projection-domain DL, researchers utilized the full-size projection data associated with the entire CT image \cite{wu2021drone, wu2022deep} or small projection patches unrelated to the CT image patches \cite{lee2018deep, dong2019sinogram, hu2020hybrid}. Recently, dual-domain DLs \cite{lin2019dudonet, zheng2020dual, wu2021drone, wang2021indudonet, han2022end} is being actively researched to achieve better performance than uni-modality DLs. However, their study merely connected the projection domain and the image domain DLs sequentially.

The study proposes a novel dual-domain DL framework designed to reconstruct missing projection data and remove remaining CT image artifacts. Specifically, the proposed method improves performance by exploiting robust mathematical properties to achieve low rankness \cite{ye2018deep}. The research reveals that the low-rank characteristics is closely related to the bowtie support of the projection data within the Fourier domain \cite{rattey1981sampling}. Furthermore, this study shows that performance is improved by controlling the low rankness through the bowtie support for hierarchical decomposed projection data associated with the CT image-patch. In particular, this paper demonstrates that the proposed method performs better when trained with higher-order hierarchical decomposed projection data. The major contributions of this paper are as follows:

\begin{itemize}
\item {The study provides evidence that the utilization of bowtie support in projection data within the Fourier domain serves as a powerful mathematical clue to improve the performance of projection-domain DL based on deep convolutional framelets (DCF) theory \cite{ye2018deep}.}
\item {The study proposes a novel dual-domain DL framework that satisfies low rankness because it is trained with high-order hierarchically decomposed projection data using a narrow bowtie support within the Fourier domain, designed to reconstruct missing projection data.}
\item {The study demonstrates that using higher-order hierarchically decomposed projection data associated with CT image-patches shows better performance when training projection-domain DLs.}
\end{itemize}

\section{Related Works}
\label{sec:2}
\subsection{Image-domain Deep Learning Approaches}
\label{sec:2.1}
The image-domain DL approach, as illustrated in Figure \ref{fig:sparse_view}(a) acts as an image artifact removers, performing a nonlinear mapping from the sparse-view CT image to the full-view CT image. This approach is recognized as a post-processing technique because the DL is directly applied to corrupted CT images reconstructed from incomplete measurements. Commonly, to perform image-domain DL approaches, many researchers 
\cite{lee2020sparse, zhang2020redaep, xie2018artifact, han2018framing} 
follow the flowcharts shown in Figure \ref{fig:methods}(a). In terms of the network architecture, Xie \etal \cite{xie2018artifact} used the GoogLeNet structure to remove streaking artifacts. Han \etal \cite{han2018framing} developed a framing U-Net to preserve high-frequency features. Lee \etal \cite{lee2020sparse} combined multi-level wavelet operations with a U-Net architecture. From the perspective of the loss function, Zhang \etal \cite{zhang2020redaep} used $L^p~(0<p<2)$ regression loss rather than $L^2$ regression loss to preserve texture details. Thaler \etal \cite{thaler2018sparse} employed Wasserstein loss with a generative adversarial network (GAN) to improve sharpness and retain structural information in reconstructed images.

\subsection{Projection-domain Deep Learning Approaches}
\label{sec:2.2}
Unlike the image-domain DL approaches, the projection-domain DL approach is a pre-prossessing technique because it reconstructs incomplete measurements rather than corrupted images, as shown in Figure \ref{fig:sparse_view}(b). Dong \etal \cite{dong2019sinogram} used a U-Net trained with small patches of linear-interpolated projection data to synthesize a full-view projection. Lee \etal \cite{lee2018deep} proposed a fully convolutional U-Net by replacing the pooling layer with a convolutional layer and trained the network with small patches of interpolated projection data. A limitation of previous approaches is that, although the goal of the network is to reconstruct high-quality CT images, there is no relationship between the small measurement patches used to train the network and the patches in the CT images.

\subsection{Dual-domain Deep Learning Approaches}
\label{sec:2.3}
The dual-domain DL framework is a hybrid network that sequentially connects the projection-domain DL and the image-domain DL. Particularly, the dual-domain DL was developed to marge the advantage of both DLs simultaneously. Zheng \etal \cite{zheng2020dual} used two sequential U-Nets to reconstruct sparse-view measurements and enhance CT images. We \etal \cite{wu2021drone} used a discriminator to match the CT image distribution between generated images from sparse-view CT and corresponding labels. However, since the approaches are only a structured and sequential network connection, it is difficult to find suitable data processing and theoretical information flow principles.

To address the limitation of previous DL-based approaches, the study proposes a novel dual-domain DL framework designed to reconstruct incomplete projection patches associated with CT image patches, as shown in Figure \ref{fig:methods}(b). The proposed method consists of 4 parts: (1) hierarchical decomposition of filtered projection data related to CT image-patches, (2) projection-domain DL model to reconstruct a decomposed full-view projection for the reconstruction of CT image-patches, (3) image-domain DL to correct remaining image noise and artifacts, and (4) composition of the entire CT image using the reconstructed CT image-patches. 
The hierarchical decomposition part establishes an explicit relationship between projection-patches and CT image-patches and provides a mathematical foundation for achieving performance improvement due to the low-rank property through Fourier-domain support constraints. 




\section{Theory}
\label{sec:3}
\subsection{Deep Convolutional Framelets}
\label{sec:3.1}
DCF theory
\cite{ye2018deep} 
has established a mathematical connection between traditional signal processing and deep learning. This connection originates from the Hankel matrix approaches 
\cite{jin2016general}, 
leading to the formulation of a regression problem with a constraint defined by a low-rank Hankel structured matrix, as outlined below:
\begin{eqnarray}
\label{eq:aloha}
\arg \min_{\bar{f} \in \mathbb{R}^n}&~{|| {f - \bar{f}} ||^2} \\ \nonumber
\rm{subject~to}&~\small{\textrm{RANK}}\mathbb{H}_d(\bar{f}) = r < d,
\end{eqnarray}
where $f\in\mathbb{R}^{n}$ and $\bar{f}\in\mathbb{R}^{n}$ represent a label image and predicted image, respectively. $n$ is a length of the signal, $r$ denotes a rank of the Hankel structured matrix $\mathbb{H}_d(\bar{f}) \in \mathbb{R}^{n \times d}$, and $d$ is a matrix pencil parameter. Notably, the rank of the Hankel structured matrix $\small{\textrm{RANK}}\mathbb{H}_d(\bar{f})$ is determined by the number of non-zero components in the Fourier domain of the solution $\mathscr{F}(\bar{f})$:
\begin{eqnarray}
\label{eq:num_rank}
\small{\textrm{RANK}}\mathbb{H}_d(\bar{f}) = \small{\textrm{COUNT}}\left(\mathscr{F}(\bar{f}) \neq 0\right).
\end{eqnarray}
If a feasible solution $\bar{f}$ exists, the singular value decompostion (SVD) of its Hankel structured matrix $\mathbb{H}_d(\bar{f})$ can be expressed as $\textrm{SVD}(\mathbb{H}_d(\bar{f})) = U \Sigma V^T$, where $U\in\mathbb{R}^{n \times r}$ and $V\in\mathbb{R}^{d \times r}$ denote the left and right singular vector bases matrices, respectively, and $\Sigma = (\sigma) \in \mathbb{R}^{r \times r}$ represents the diagonal matrix of singular values. In this context, we consider two pairs of matrices $\Phi, \tilde{\Phi} \in \mathbb{R}^{n \times n}$ and $\Psi, \tilde{\Psi} \in \mathbb{R}^{d \times r}$, which satisfy the following conditions:
\begin{eqnarray}
\label{eq:frames}
(a)~\tilde{\Phi} \Phi^T = I_{n \times n},~~~~~(b)~\Psi \tilde{\Psi}^T = P_{R(V)},
\end{eqnarray}
where $R(V)$ denotes a range space of $V$, and $P_{R(V)}$ represents a projection onto $R(V)$. Using Eq. \ref{eq:frames}, we can formulate an equality of the Hankel structured matrix $\mathbb{H}_d(\bar{f})$, given by:
\begin{eqnarray}
\label{eq:hankel}
\mathbb{H}_d(\bar{f}) = \tilde{\Phi} \Phi^T \mathbb{H}_d(\bar{f}) \Psi \tilde{\Psi}^T.
\end{eqnarray}
From Eq. \ref{eq:hankel}, we can establish a space $\mathcal{F}_r$ collecting feasible images $\bar{f}$, as follows:
\begin{eqnarray}
\label{eq:set_h}
\mathcal{F}_r = \left\{ {\bar{f}\in\mathbb{R}^n \bigg| \bar{f} = \left( {\tilde{\Phi}C} \right) \circledast \nu(\tilde{\Psi}), C=\Phi^T(\bar{f}\circledast \bar{\Psi}) } \right\},
\end{eqnarray}
where $\bar{\Psi}$ and $\nu(\tilde{\Psi})$ denote encoder- and decoder-layer convolutional filters, respectively.
The regression problem initially in Eq. \ref{eq:aloha} can be reformulated using the space $\mathcal{F}_r$ as follows: 
\begin{eqnarray}
\label{eq:aloha_w_H}
\arg \min_{\bar{f} \in \mathcal{F}_r}{|| {f - \bar{f}} ||^2},
\end{eqnarray}
which can be expressed by optimizing kernels $(\Psi, \tilde{\Psi})$ of neural network $\mathcal{Q}$ as follows:
\begin{eqnarray}
\label{eq:op_kernel}
\arg \min_{(\Psi, \tilde{\Psi})}{|| {f - \mathcal{Q}(q; \Psi, \tilde{\Psi})} ||^2},
\end{eqnarray}
where $q$ is a noisy image.
The neural network $\mathcal{Q}$ can be trained with extensive datasets $\{ (q^{(i)}, f^{(i)}) \}_{i=1}^N$ to learn the kernels $(\Psi, \tilde{\Psi})$ that represent $\small{\textrm{RANK}}\mathbb{H}_d(\bar{f}^{(i)}) \leq r_{\textrm{max}}$, where $r_{\textrm{max}}$ is the largest rank of the Hankel structured matrix $\mathbb{H}_d(\bar{f}^{(i)})$ among the datasets, and $d$ is redefined by the convolutional filter length, 
as described by:
\begin{eqnarray}
\label{eq:op_net}
\arg \min_{(\Psi, \tilde{\Psi})}\sum_{i=1}^{N}{|| {f^{(i)} - \mathcal{Q}(q^{(i)}; \Psi, \tilde{\Psi})} ||^2}.
\end{eqnarray}

From Eq. \ref{eq:aloha}, the largest rank $r_{\textrm{max}}$ of Hankel structured matrix $\small{\textrm{RANK}}\mathbb{H}_d(\bar{f})$ is bounded by the convolutional filter length $d$. To satisfy the low-rank property, the filter length $d$ can be increased until the signal length $n$, but a network architecture with long filter length $d \simeq n$ is difficult to utilized due to computational resources and efficiency. Therefore, to improve the performance when the network architecture $\mathcal{Q}$ is fixed, $r_{\textrm{max}}$ should be reduced. An easy way to reduce  $r_{\textrm{max}}$ is to reduce the non-zero components of the signal $\bar{f}$ in the Fourier domain based on Eq. \ref{eq:num_rank}.

\subsection{Bowtie Support in the Fourier domain}
\label{sec:3.2}

In the following, the paper first delineates the Radon transform, denoted by $\mathscr{R}$, and subsequently extend it to describe a bowtie support of measurements in the Fourier domain. Let $\theta$ represent a vector on the unit sphere $\mathbb{S} \in \mathbb{R}^2$ . The set of orthogonal vectors, denoted by $\theta^{\perp}$, is characterized as:
\begin{eqnarray}
	\mathbf{\theta}^\perp = \{\mathbf{v} \in \mathbb{R}^2 : \mathbf{v} \cdot \mathbf{\theta} = 0\},
\end{eqnarray}
where $\cdot$ represents an inner product. If an image is defined by $f(\mathbf{x})$ for $\mathbf{x} \in \mathbb{R}^2$, the Radon transform $\mathscr{R}$ of the image $f$ can be expressed as follows: 
\begin{eqnarray}
\label{eq:radon}
 	\mathscr{R}f(\mathbf{\theta}, u) &=& p_f(\mathbf{\theta}, u) \\ \nonumber
&=& \int_{\mathbf{\theta}^\perp} d\mathbf{v}~{f(\mathbf{v} + u\mathbf{\theta})},
\end{eqnarray}
where $u\in \mathbb{R}$ and $\mathbf{\theta}\in\mathbb{S}$. In order to evaluate the maximum rank $r_{\textrm{max}}$ of the Hankel structured matrix corresponding to the measurements $p_f$ 
as defined in Eq. \ref{eq:num_rank}, the 2D Fourier transform $\mathscr{F}_{2D}$ is applied to Eq. \ref{eq:radon}, as follows:
\begin{eqnarray}
\label{eq:ft_radon}
 	\mathscr{F}_{2D}p_f(\mathbf{\theta}, u) &=& P_f(\omega_\theta, \omega_u) \\ \nonumber
  &=& \int_{-\infty}^{\infty}\int_{-\infty}^{\infty}d\theta du ~{p_f(\mathbf{\theta}, u)e^{-j(\omega_\theta \theta+\omega_u u)}}.
\end{eqnarray}

\begin{figure}[!t]
    \centering
    \includegraphics[width=0.8\textwidth]{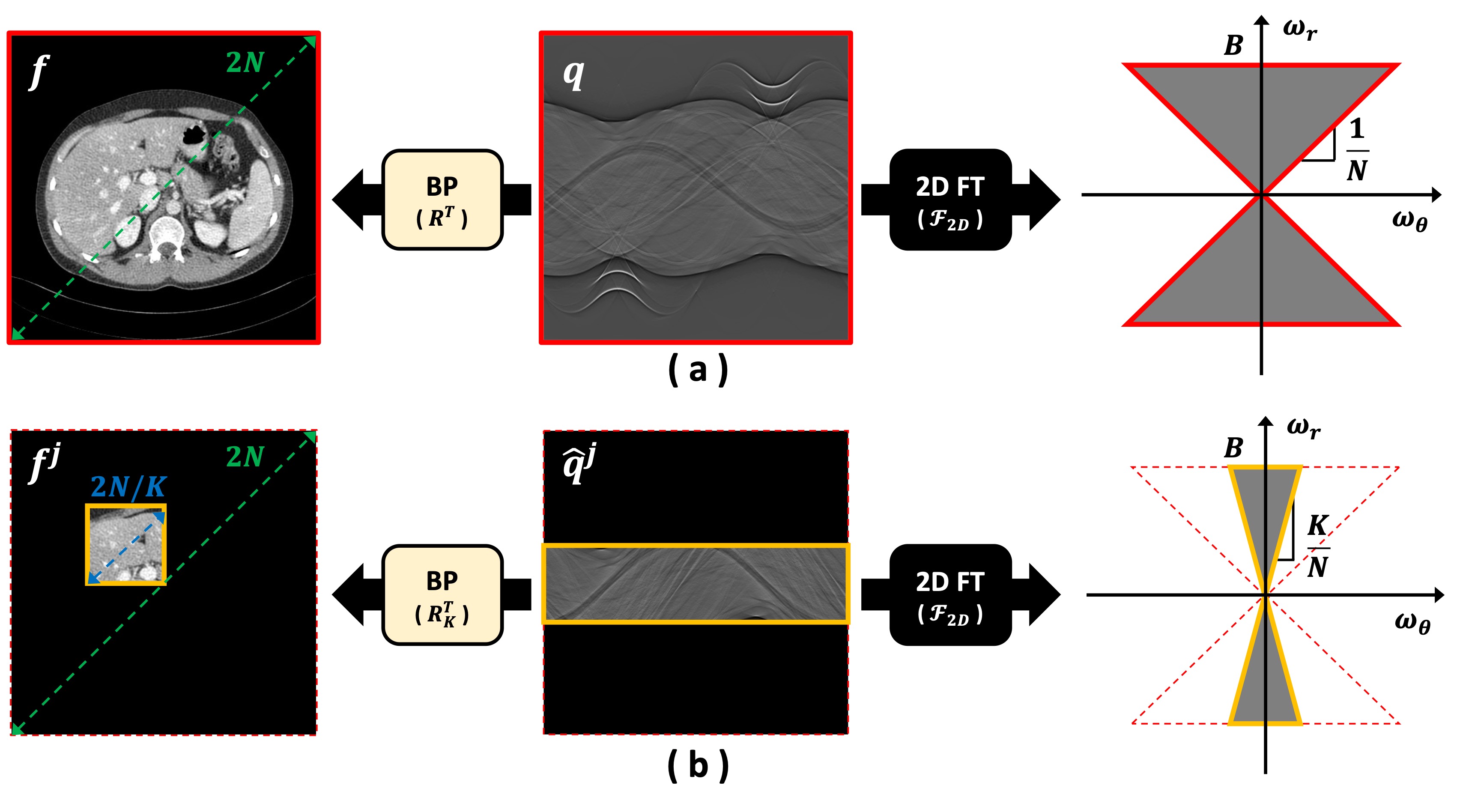}
    \caption{Bowtie support in the Fourier domain according to (a) K = 1 and (b) K = 3. Here, $N$ is a radius of object, $K$ is a decomposition level, and $B$ is a bandlimit.}
    \label{fig:bowtie}
\end{figure}

Thanks to Rattey \etal \cite{rattey1981sampling}, the measurement $P_f(\omega_\theta, \omega_u)$ is bounded by a bowtie support in the Fourier domain, as shown in Figure \ref{fig:bowtie}(a). A slope of the bowtie support is determined by $\frac{1}{N}$ , where $N$ denotes a radius of the entire CT image $f$. Consequently, the rank $r$ of Hankel structured matrix $\small{\textrm{RANK}}\mathbb{H}_d(p_f)$ can be determined as the area of bowtie support within the Fourier domain. The filtered measurements $q = \mathcal{F}(p_f)$ is equivalent to an element-wise weighted measurements in the Fourier domain. Therefore, they have the same bowtie support and satisfy the same rank $r$.

\begin{figure}[!t]
    \centering
    \includegraphics[width=0.7\textwidth]{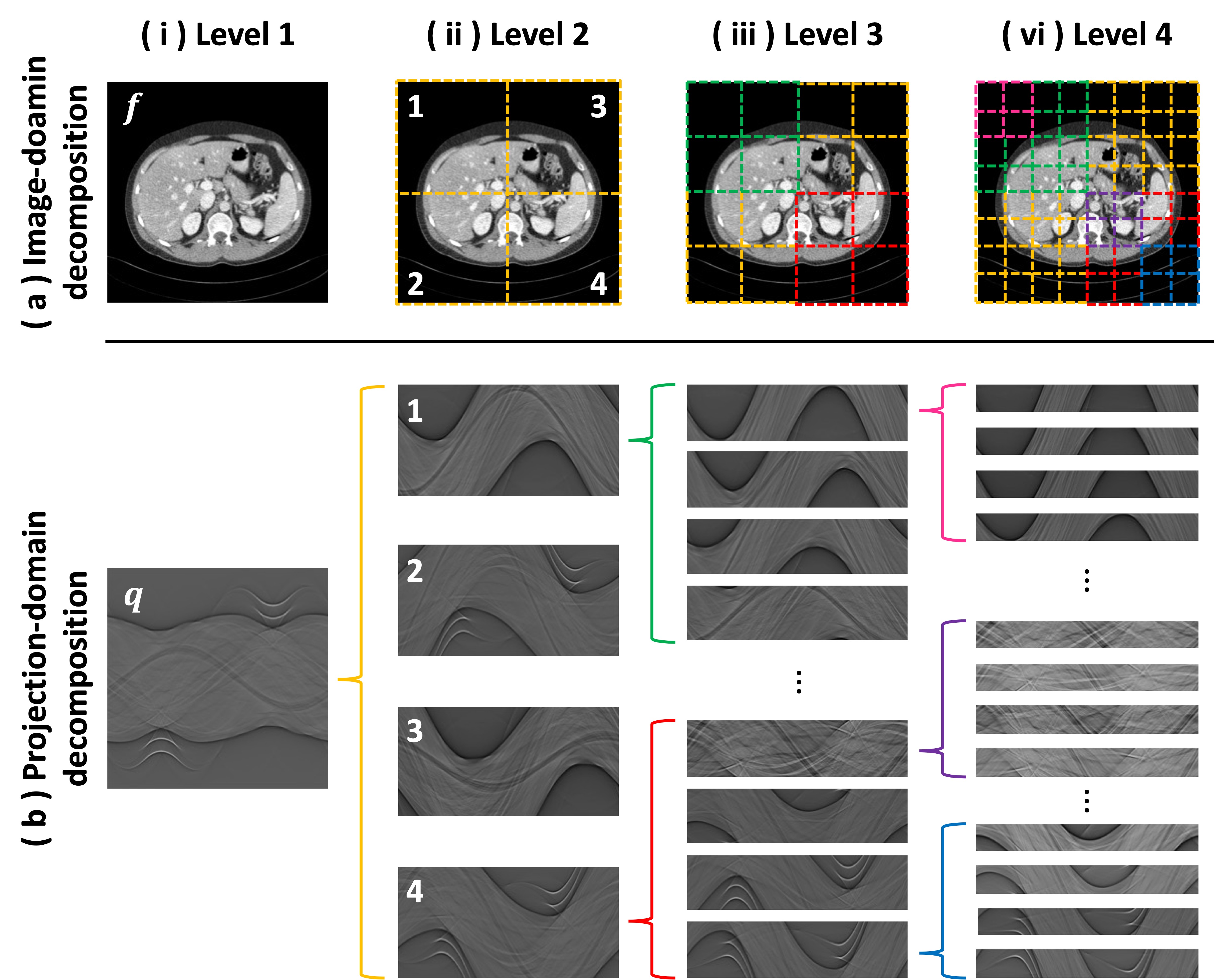}
    \caption{Hierarchical decomposition concept for decomposition levels of (a) image-domain and (b) projection-domain. }
    \label{fig:decomposition}
\end{figure}

\subsection{Hierarchical Decomposition of Measurements}
\label{sec:3.3}

From the analysis in the previous sections \ref{sec:3.2}, it has been observed that a measurement collected by the Radon transform $\mathscr{R}$ exists within the bowtie support in the Fourier domain, as grounded on the work by \cite{rattey1981sampling}. Additionally, the support area ($2B^2N$) has been found to correspond with the rank $r$ of the Hankel structured matrix of the measurement $\mathbb{H}_d({q})$, as elucidated by the DCF theory \cite{ye2018deep}. 
When a network architecture is fixed, the improvement of network performance is closely related to how well it achieves the lower rankness of the Hankel structured matrix. In other words, a smaller support area generally indicates better performance. To accomplish a reduction of the support area in the Fourier domain, a hierarchical decomposition algorithm \cite{basus2000n2log2n} is applied to the measurement, as illustrated in Figure \ref{fig:bowtie}(b). By reducing the image size $N$ by a decomposition level $K$, the area ($(2B^2N)/K$) can also be diminished by the same proportion. Therefore, the low-rank property is achieved through a decomposed projection associated with the image-patch. The concept of the decomposition is illustrated in Figure \ref{fig:decomposition}, which shows both (a) image-patches and (b) projection-patches in accordance with the decomposition level $K$. The decomposition process of projection-patches $\mathcal{D}_{prj}^{K}$ is described in the Algorithm \ref{alg:decomposition}. 



\begin{algorithm} [t!]
\caption{Hierarchical measurement decomposition}
\label{alg:decomposition}
\begin{algorithmic}[1]
\Require $\textit{\textbf{q}}, N_x, N_y, N_{dct}, N_{view}, K$
\State $N_{K} \gets 2^{(K-1)}$
\State $(N_{x}^{K},~N_{y}^{K},~N_{dct}^{K}) \gets (\frac{N_{x}}{N_{K}},~\frac{N_{y}}{N_{K}},~\frac{N_{dct}}{N_{K}})$

\State $\textit{\textbf{x}}_C \gets \text{linspace}(-(\frac{N_x}{2} - \frac{N_x}{2^K}), (\frac{N_x}{2} - \frac{N_x}{2^K}), N_{x}^{K})$~{{\color{teal}{\# X-axis center position of image patch}}}
\State $\textit{\textbf{y}}_C \gets \text{linspace}(-(\frac{N_y}{2} - \frac{N_y}{2^K}), (\frac{N_y}{2} - \frac{N_y}{2^K}), N_{y}^{K})$~{{\color{teal}{\# Y-axis center position of image patch}}}

\State $\theta \gets \text{linspace}(0, 2\pi, N_{view})$
\State $\bf{M}_{rot} \gets [\cos(\theta), \sin(\theta)]$~~~~~~~~~~~~~~~~~~~~~~~~~~~{{\color{teal}{\# Rotation matrix along view angle}}}

\State $\textit{\textbf{q}}_{K} \gets \text{zeros}(N_{K}^2,~N_{view},~N_{dct}^{K})$ 
\For{$j=0:N_{K}$}
\For{$i=0:N_{K}$}
\State $\textit{\textbf{dct}}_C \gets \bf{M}_{rot}\cdot[\textit{\textbf{y}}_C$$(i)$, $\textit{\textbf{x}}_C$$(j)]$~~~~~{{\color{teal}{\# Detector center position for image patch}}}
\State $\tilde{\textit{\textbf{q}}}_{K} \gets \text{Align}(\textit{\textbf{q}}, \textit{\textbf{dct}}_C)$~~~~~~~~~~~~~~~{{\color{teal}{\# Align $q$ along detector center line}}}
\State $\hat{\textit{\textbf{q}}}_{K} \gets \text{Extract}(\tilde{\textit{\textbf{q}}}_{K}, N_{dct}^{K})$~~~~~~~~~~~{{\color{teal}{\# Extract aligned patch $\tilde{q}$}}}
\State $k \gets N_{y}^{K} \times j + i$
\State $\textit{\textbf{q}}_{K}(k) \gets \hat{\textit{\textbf{q}}}_{K}$~~~~~~~~~~~~~~~{{\color{teal}{\# Collect decomposed measurement $\hat{q}$ into level $K$}}}
\EndFor
\EndFor \\
\Return $\textit{\textbf{q}}_{K}$
\end{algorithmic}
\end{algorithm}

\section{Main Contributions}
\label{sec:4}
In Section \ref{sec:3}, the paper established that the DL performance is closely related to a low rankness of a Hankel structured matrix of data in the Fourier domain, as per the DCF theory \cite{ye2018deep}. The projection data also exhibits bowtie support in the Fourier domain \cite{rattey1981sampling} and can be hierarchically decomposed into forms with narrow bowtie support \cite{basus2000n2log2n}. Thanks to the DCF theory \cite{ye2018deep} and the hierarchical decomposition method \cite{basus2000n2log2n}, this study can achieve lower rankness through higher hierarchical decomposition of the measurements.

Based on two mathematical clues, this study proposes a novel hierarchical decomposed dual-domain DL (PI-Net; $\mathcal{Q}_{prj^1}^K$, $\mathcal{Q}_{img^2}^K$) as shown in Figure \ref{fig:methods}(b). To enforce the low-rank property, a projection-domain hierarchical decomposition $\mathcal{D}_{prj}^K$ (see Figure \ref{fig:decomposition}(b)) is applied to the measurements, as illustrated in Figure \ref{fig:methods}(b)(i). Next, the projection-domain network (P-Net; $\mathcal{Q}_{prj}^K$) is trained by the decomposed measurements with various DS factors $\mathcal{S}$ applied simultaneously, as shown in Figure \ref{fig:methods}(b)(ii). Once the P-Net $\mathcal{Q}_{prj}^K$ is fixed, the maximum available rank $r_{max}$ is also determined. However, due to various DS factors $\mathcal{S}$, the required rank may increase as the number of training datasets increased. Here, thanks to the projection-domain hierarchical decomposition $\mathcal{D}_{prj}^K$, the P-Net $\mathcal{Q}_{prj}^K$ can be well-trained with the decomposed measurements because the narrow bowtie support of decomposed measurements in the Fourier domain reduces the required ranks. As shown in Figure \ref{fig:methods}(b)(iii), the image-domain network (I-Net; $\mathcal{Q}_{img}^K$) is connected as an unrolled scheme to correct remaining artifacts after P-Net $\mathcal{Q}_{prj}^K$. Finally, an image-domain composition $\mathcal{C}_{img}^K$ is applied to the image-patches reconstructed from the I-Net $\mathcal{Q}_{img}^K$ in order to convert it into an entire CT image. In the paper, the decomposition level $K=5$ is used. To compare with the proposed PI-Net in Figure \ref{fig:methods}(b), a decomposed image-domain DL (II-Net; $\mathcal{Q}_{img^1}^K$, $\mathcal{Q}_{img^2}^K$) consisting of two-times unrolled I-Net $\mathcal{Q}_{img}^K$ was used, as shown in Figure \ref{fig:methods}(a).

\begin{algorithm} [t!]
\begin{algorithmic}[1]
\Require $p, \mathcal{M}_\mathcal{S}$ 
\State $p_\mathcal{S} \gets \mathcal{M}_\mathcal{S} \odot p$ ~~~~~~~~~~~~~~~~~~~~~~~~~{{\color{teal}{\# Sparse-view measurement}}}
\State $f_\mathcal{S} \gets \mathcal{R}^T_{\mathcal{S}} (\mathcal{F}(q_\mathcal{S}))$ ~~~~~~~~~~~~~~~~~~~~~{{\color{teal}{\# Input data of $\mathcal{Q}_{img^1}$}}}
\State $\hat{p}_\mathcal{S} \gets \mathcal{R}(f_\mathcal{S})$ ~~~~~~~~~~~~~~~~~~~~~~~~~~~{{\color{teal}{\# Interpolated measurement}}}
\State $\bar{p}_\mathcal{S} \gets (1 - \mathcal{M}_\mathcal{S}) \odot \hat{p}_\mathcal{S} + \mathcal{M}_\mathcal{S} \odot p$ ~{{\color{teal}{\# Data-consistency regularization}}}

\State $q_\mathcal{S} \gets \mathcal{F}(\bar{p}_\mathcal{S})$   ~~~~~~~~~~~~~~~~~~~~~~~~~~~{{\color{teal}{\# Input data of $\mathcal{Q}_{prj^1}$}}}
\\
\Return $f_\mathcal{S}, q_\mathcal{S}$
\end{algorithmic}
\caption{Input data generation for $\mathcal{Q}_{img^1}$ and $\mathcal{Q}_{prj^1}$}
\label{alg:data_gen}
\end{algorithm}

\section{Experiments}
\label{sec:4}
\subsection{Datasets}
\label{sec:4.4}


For this study, ten subject datasets were sourced from the American Association of Physicists in Medicine (AAPM) Low-Dose CT Grand Challenge \cite{mccollough2016tu}. These datasets were utilized in the following: among the ten subjects, nine were allocated to training and validation. Specifically, eight subjects comprising 4,006 slices were used for training, and one subject with 254 slices was designated for validation. The remaining subject, containing 486 slices, was employed as the test dataset. Although the parallel beam CT geometry was used in this experiment, the measurements collected from fan beam CT can also be utilized by rebinning to the form of parallel beam measurements. The image size ($N_x,~N_y$) is ($512,~512$), with a pixel resolution of $1~mm^2$. The number of views $N_{view}$ is $768$, and the rotation range for the X-ray source is $[0^\circ,~360^\circ)$. The number of detectors $N_{dct}$ is $768$, and the detector pitch is $1~mm$. 
Additionally, the decomposition levels $K$ were set at 1, 2, 3, 4 and 5. 
Each level expands the number of dataset by a factor $J = 2^{(2\times(K-1))}$ of 1, 4, 16, 64, and 256, respectively, reduces the detector size and the image size according to $N_{dct}^K = N_{dct} / 2^{K-1}$ and $N_{x,y}^K = N_{x,y} / 2^{K-1}$. 
Downsampling (DS) ratios used to synthesize the sparse-view measurement were [2, 3, 4, 6, 8, 12], and each sparse-view corresponding to the DS ratio was [384, 256, 192, 128, 96, 64] views.
In particular, to match the size of the measurements input to the DL without unintentional interference affecting performance, Algorithm \ref{alg:data_gen} is applied to the downsampled measurement to interpolate them to the full-view size. The symbolic details are described in Table \ref{tbl:obj}(f). When sparse-view data is generated for training the projection-domain DL, a DS factor $\mathcal{S}$ is randomly selected from among the DS ratios, and the synthetic undersampled measurements, applied to given DS factor $\mathcal{S}$, are generated using Algorithm \ref{alg:data_gen}.


\subsection{Architectures}
\label{sec:4.2}

\begin{figure}[!t]
    \centering
    \includegraphics[width=0.9\textwidth]{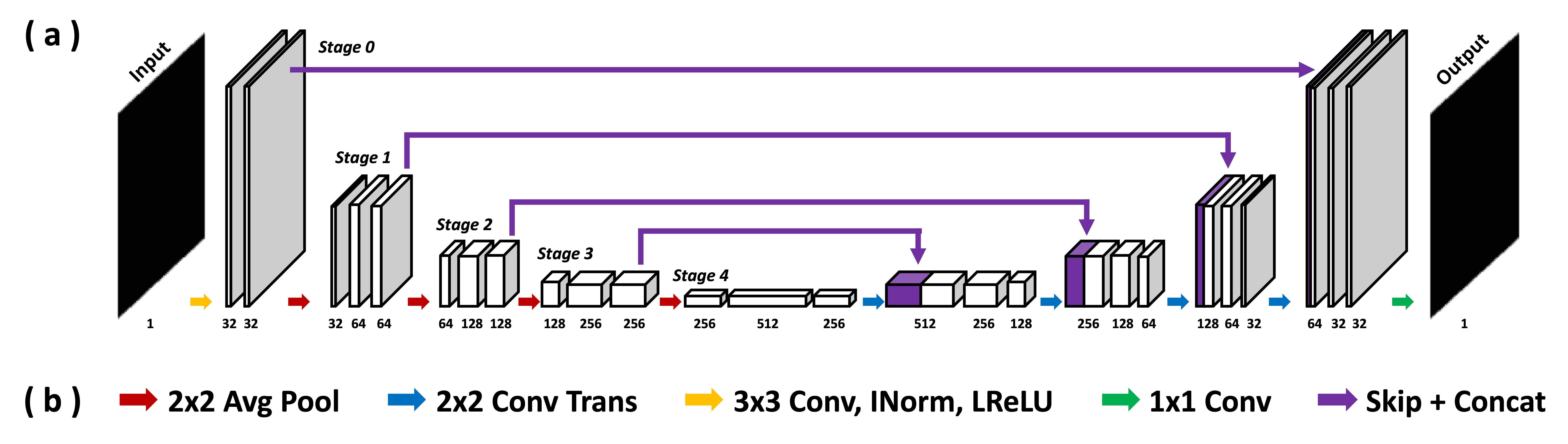}
    \caption{(a) Backbone based on the standard U-Net architecture. (b) Layer modules used in (a). }
    \label{fig:network}
\end{figure}
To evaluate the DL performance across data domains, including image-domain and projection-domain, two types of DL frameworks are used, as shown in Figure \ref{fig:methods}: (a) II-Net consisting of two-times unrolled I-Net $\mathcal{Q}_{img}^K$, and (b) PI-Net sequentially composed of P-Net $\mathcal{Q}_{prj}^K$ and I-Net $\mathcal{Q}_{img}^K$. 
Figure \ref{fig:network} shows a backbone architecture and layer modules used. P-Net $\mathcal{Q}_{prj}^K$ and I-Net $\mathcal{Q}_{img}^K$ use the same backbone architecture, as shown in Figure \ref{fig:network}(a). A basic layer block consists of a $3 \times 3$ convolutional layer ($3\times3$ Conv), instance normalization (INorm), and a leaky rectified linear unit (LReLU) as illustrated by the yellow arrow in Figure \ref{fig:network}(b). The basic layer block is present between all blocks in Figure \ref{fig:network}(a), but the yellow arrow has been omitted for visibility. The backbone network has 7,764,049 trainable parameters, and both II-Net and PI-Net are composed of two backbone networks, resulting in a total of 15,528,098 (=7,764,049 + 7,764,049) parameters.

\subsection{Training}
\label{sec:4.3}

\begin{table*}[t!]
    \centering
    \begin{adjustbox}{width=1.0\textwidth}
    \begin{tabular}{ l l l l l } 
    \hline
    
    &&&& \\ 
    

    &&&& \\ 
    (a) $\mathscr{L}_{II-Net}^{K}$ & \multicolumn{4}{l}{$\arg \min_{(\mathcal{Q}_{img^1}^K, \mathcal{Q}_{img^2}^K)}{\sum_{i=1}^{I}\sum_{j=1}^{J} \underbrace{\Vert f_K^{i,j} - (f_{K, \mathcal{S}}^{i,j} - \mathcal{Q}_{img^1}^K (f_{K, \mathcal{S}}^{i, j})) \Vert ^2}_{\text{( i ) 1st phase network}} } + {\sum_{i=1}^{I}\sum_{j=1}^{J} \underbrace{\Vert f_K^{i,j} - (\bar{f}_{K, img^1}^{i, j} - \mathcal{Q}_{img^2}^K (\bar{f}_{K, img^1}^{i, j})) \Vert ^2}_{\text{( ii ) 2nd phase network}}}$,} \\ 
    &&&& \\ 
    & \multicolumn{4}{l}{where $\bar{f}_{K, img^1}^{i, j} = f_{K, \mathcal{S}}^{i,j} - \mathcal{Q}_{img^1}^K (f_{K, \mathcal{S}}^{i, j})$,~~~$f_{K, \mathcal{S}}^{i, j} = \mathcal{D}^K_{img}(f_\mathcal{S}^i)[j]$ and $f_{K, \mathcal{S}} = \mathcal{R}^T_K(q_{K, \mathcal{S}})$.} \\ 
    &&&& \\ 
    \hline

    &&&& \\ 
    (b) $\mathscr{L}_{PI-Net}^{K} $ & \multicolumn{4}{l}{$\arg \min_{(\mathcal{Q}_{prj^1}^K, \mathcal{Q}_{img^2}^K)} {\sum_{i=1}^{I}\sum_{j=1}^{J} \underbrace{\Vert f_K^{i,j} - \mathscr{R}^T_K(\mathcal{M}_\mathcal{S} \odot q_{K, \mathcal{S}}^{i, j} + (1 - \mathcal{M}_\mathcal{S}) \odot \mathcal{Q}_{prj^1}^K(q_{K, \mathcal{S}}^{i, j}) ) \Vert ^2}_{\text{( i ) 1st phase network}} } + {\sum_{i=1}^{I}\sum_{j=1}^{J} \underbrace{\Vert f_K^{i,j} - ( \bar{f}_{K, prj}^{i, j} - \mathcal{Q}_{img^2}^K (\bar{f}_{K, prj}^{i, j})) \Vert ^2}_{\text{( ii ) 2nd phase network}}}$,} \\ 
    &&&& \\ 
    & \multicolumn{4}{l}{where $\bar{f}_{K, prj^1}^{i, j} = \mathscr{R}^T_K(\mathcal{M}_\mathcal{S} \odot q_{K, \mathcal{S}}^{i, j} + (1 - \mathcal{M}_\mathcal{S}) \odot \mathcal{Q}_{prj^1}^K(q_{K, \mathcal{S}}^{i, j}) )$ and $q_{K, \mathcal{S}}^{i, j} = \mathcal{D}^K_{prj}(q_\mathcal{S}^i)[j]$.} \\ 
    &&&& \\ 
    \hline
    
    &&&& \\ 
    
    (c) $\mathscr{L}_{TV}$ & \multicolumn{4}{l}{$\arg \min_{f} \frac{1}{2} \Vert p_{\mathcal{S}} - \mathscr{R}(f) \Vert _2^2 + \lambda TV(f)$,} \\ 
    &&&& \\ 
    & \multicolumn{4}{l}{where $\lambda$ denotes an weight parameter, which balances the fidelity term and the TV regularization term.} \\ 
    &&&& \\ 
    \hline
    &&&& \\ 
    (d) Symbols & $\odot$ & Hadamard product & $K$ & Decomposition level ($K$ is omitted for visibility when $K$=1.) \\
                & $I$ & Number of datasets & $J$ & Number of decomposed data ($J= 2^{(2\times(K-1))}$) \\
                & $p$ & Full-view projection data & $\mathcal{M}_\mathcal{S}$ & Sparse-view projection mask \\
                & $p_{\mathcal{S}}$ & Sparse-view projection data & $q_{K, \mathcal{S}}$ & Sparse-view filtered projection data decomposed into $K$ level \\
                & $f_{K}$ & Full-view CT image decomposed into level $K$  & $f_{K, \mathcal{S}}$ & Sparse-view CT image decomposed into level $K$ \\
                & $\bar{f}_{K, prj}$ & Reconstructed CT image from projection-domain DL $\mathcal{Q}_{prj}^K$ & $\bar{f}_{K, img}$ & Reconstructed CT image from image-domain DL $\mathcal{Q}_{img}^K$ \\
                & $\mathcal{R}$ & Projection operation & $\mathcal{R}^T_K$ & Backprojection operation for $K$ level decomposition \\
                & $TV$ & Total variation operation &  $\mathcal{F}$ & Filtration operation \\
                & $\mathcal{Q}_{prj}^K$ & Projection-domain DL for $K$ level decomposition & $\mathcal{Q}_{img}^K$ & Image-domain DL for $K$ level decomposition \\
                & $\mathcal{D}_{prj}^K$ & Projection-domain decomposition into level $K$ & $\mathcal{D}_{img}^K$ & Image-domain decomposition into level $K$ \\
    &&&& \\ 
    \hline

    \end{tabular}
    \end{adjustbox}
    \caption{Objective functions for (a) II-Net, (b) PI-Net and (c) MBIR with TV regularization methods. (d) Various symbols used.}
    \label{tbl:obj}
\end{table*}

\subsubsection{Environments}
The DL architectures were implemented using Pytorch. To calculate the loss of the PI-Net as shown in Table \ref{tbl:obj}(b), the backprojection operation $\mathcal{R}^T$ was implemented as a user-defined layer in Pytorch. Additionally, the backward propagation of the backprojection operation $\mathcal{R}^T$ could be conducted in a sequential manner using the projection operation $\mathcal{R}$. A graphic processing unit (GPU), such as NVIDIA A6000, was used to train the networks. The hyper parameters employed for training the DLs are detailed as follows: An Adam optimizer was utilized, and the initial learning rate was set to $10^{-4}$. If the validation loss did not show a decrease over five consecutive epochs, the learning rate was multiplied by 0.1. Pairs of (the number of epoch, batch size) were defined as [(100, 4), (50, 16), (50, 32), (50, 64), (50, 256)], in accordance with the decomposition levels $K$ = [1, 2, 3, 4, 5], respectively. In particular, the DL model trained with decomposition level $K=1$ was used as a pre-trained model for other decomposition levels $K$ = [2, 3, 4, 5]. 
In the perspective of the shift-invariant characteristic of convolution layer, the bias issue caused by using the pre-trained model with decomposition level $k=1$ will be minor and it would be a stable initial point. 
Three quantitative metrics were used: the normalized root mean square error (NRMSE), the peak signal to noise ratio (PSNR), and the structural similarity index measure (SSIM).

\subsubsection{Objective functions}
Table \ref{tbl:obj} shows the various objective functions used to train various DLs and MBIR method. 
The objective function of the II-Net, defined in Table \ref{tbl:obj}(a), has two terms to train (i) the 1st I-Net $\mathcal{Q}_{img^1}^K$ and (ii) the 2nd I-Net $\mathcal{Q}_{img^2}^K$, simultaneously. Similarly, the PI-Net is trained by the objective function, consisting of two terms for (i) the 1st P-Net $\mathcal{Q}_{prj^1}^K$ and (ii) the 2nd I-Net $\mathcal{Q}_{img^2}^K$, defined in Table \ref{tbl:obj}(b). 
Specifically, in order to maintain and preserve the original measured information, the I-Net $\mathcal{Q}_{img}^K$ is trained with the concept of residual learning by modifying a label image $f_K$ into a residual image $(f_K - f_{K, \mathcal{S}})$, and a data-consistency term is applied to the P-Net $\mathcal{Q}_{prj}^K$ by replacing the reconstructed measurement $\mathcal{Q}_{prj}^K(q_{K, \mathcal{S}})$ with the original measurement $q_{K, \mathcal{S}}$ at measured view position.
In addition, MBIR algorithm with TV regularization is used for the comparative conventional method, and the cost function is formulated in Table \ref{tbl:obj}(c).

\begin{figure}[t!]
    \centering
    \includegraphics[width=1.0\textwidth]{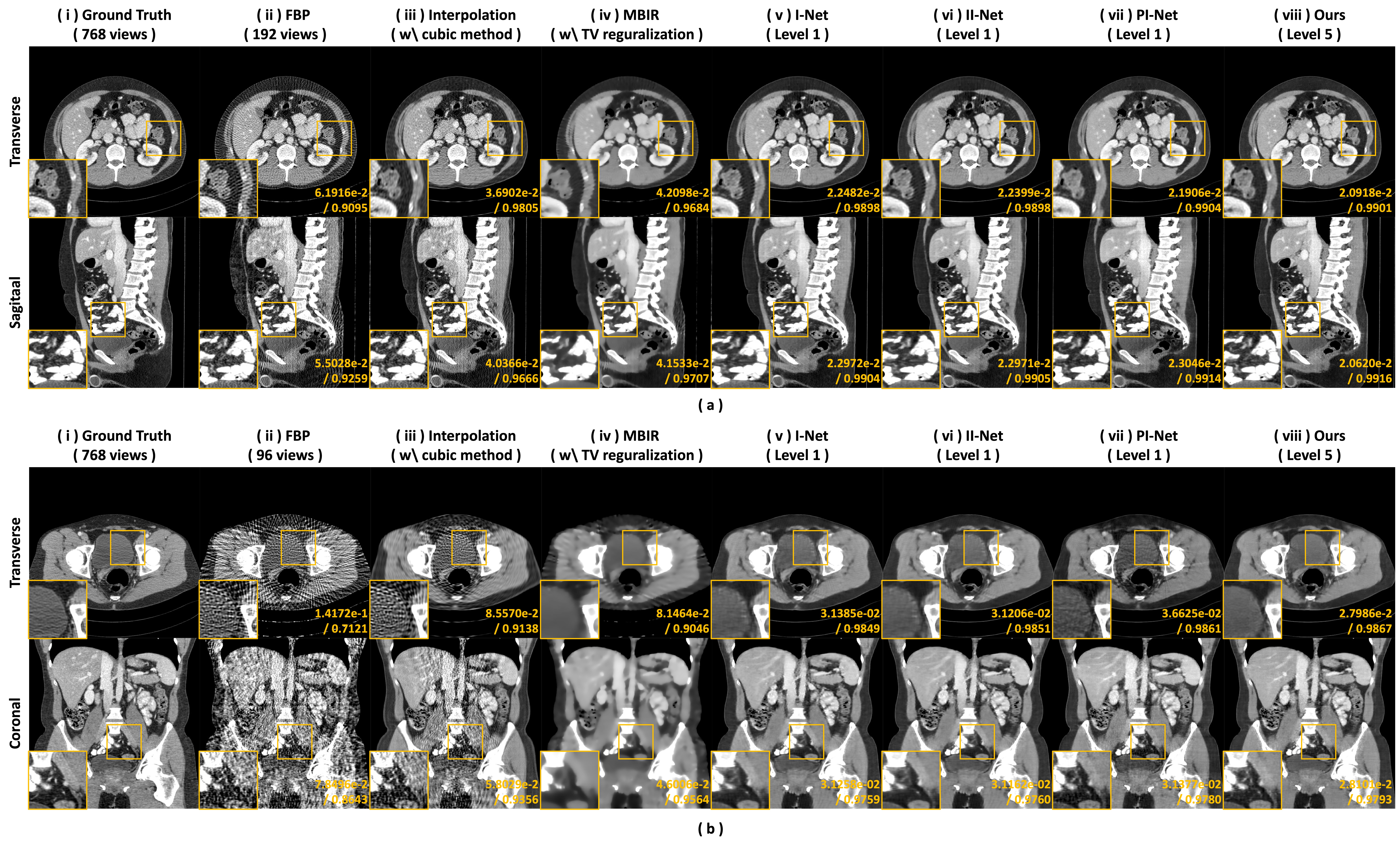}
    \caption{Various directional results from (a) 192 views (DS factor $\mathcal{S} = 4$) and (b) 96 views (DS factor $\mathcal{S} = 8$). (i) Ground truth and reconstructed images from (ii) FBP, (iii) cubic interpolation, (iv) MBIR with TV regularization, (v) I-Net, which is 1st phase network of II-Net, (vi) II-Net, (vii) PI-Net, and (viii) Proposed method. (v-vii) conventional DL approaches were trained at decomposition level $K=1$, while (viii) our method were trained  at decomposition level $K=5$. The intensity range was set to (-160, 240) [HU]. NRMSE/SSIM values are written in the corners.}
    \label{fig:result_main}
\end{figure}

\section{Results and Discussion}
\label{sec:6}
Figure \ref{fig:result_main} shows the reconstructed images from various algorithms, including the analytic method (see Figure \ref{fig:result_main}(iii)), the iterative method (see Figure \ref{fig:result_main}(iv)), and DL-based methods (see Figures \ref{fig:result_main}(v-viii)). The interpolation method in Figure \ref{fig:result_main}(iii) does not clearly remove streaking artifacts, while MBIR with a TV penalty in Figure \ref{fig:result_main}(iv) is overestimated and fails to preserve textures due to strong TV regularization. In contrast to analytic and iterative methods, the DL-based methods, including conventional DL approaches in Figures \ref{fig:result_main}(v-vii) and our method in Figure \ref{fig:result_main}(viii), show superior image quality and quantitative metrics. When comparing our method with conventional DL approaches, the images reconstructed from I-Net and II-Net in Figures \ref{fig:result_main}(v-vi) exhibit smooth textures, but the global streaking patterns persist. In Figure \ref{fig:result_main}(vii), the traditional PI-Net using decomposition level $K=1$ clearly removes the global streaking artifacts, but the image noise increases when the DS factor $\mathcal{S}$ is high, as shown in Figure \ref{fig:result_main}(b)(vii). The proposed PI-Net using decomposition level $K=5$ preserves not only the small structures but also the textures and achieves the lowest NRMSE and the best SSIM values, as illustrated in Figure \ref{fig:result_main}(viii). In particular, unlike the traditional PI-Net, our method shows stable reconstruction performance even in high DS factor environment, as shown in Figure \ref{fig:result_main}(b)(viii). 
The average quantitative metrics, including PSNR and SSIM values with respect to various numbers of views, are presented in Table \ref{tbl:result_main}. Among various algorithms, DL-based methods outperform analytic and iterative methods, and our DL method exhibits better quantitative metrics than other DL-based methods.

\begin{table*}[t!]
    \centering
    \begin{adjustbox}{width=1.0\textwidth}
    \begin{tabular}{ |c|c|c|c|c|c|c|c| } 
    \hline
    \multicolumn{1}{|c|}{\multirow{1}{*}{(a) PSNR $\uparrow$}} 
                                    & \multicolumn{1}{|c|}{\multirow{2}{*}{(i) FBP}}        & \multicolumn{1}{|c|}{\multirow{1}{*}{(ii) Interp.}}       & \multicolumn{1}{|c|}{\multirow{1}{*}{(iii) MBIR}}  
                                    & \multicolumn{1}{|c|}{\multirow{1}{*}{(iv) I-Net}}      & \multicolumn{1}{|c|}{\multirow{1}{*}{(v) II-Net}}         & \multicolumn{1}{|c|}{\multirow{1}{*}{(vi) PI-Net}}   & \multicolumn{1}{|c|}{\multirow{1}{*}{(vii) Ours}}\\
    \multicolumn{1}{|c|}{[ dB ]}    &                                                       & \multicolumn{1}{|c|}{\small( w$\backslash$ Cubic )}                  & \multicolumn{1}{|c|}{\small( w$\backslash$ TV )}                    
                                    & \multicolumn{1}{|c|}{\small( Level 1 )}               & \multicolumn{1}{|c|}{\small( Level 1 )}                   & \multicolumn{1}{|c|}{\small( Level 1 )}                   & \multicolumn{1}{|c|}{\small( Level 5 )} \\ \hline  
    \multicolumn{1}{|c|}{384 views (DS = 2)} & \multicolumn{1}{|c|}{43.8766}                         & \multicolumn{1}{|c|}{46.8335}                             & \multicolumn{1}{|c|}{45.3438}                             
                                    & \multicolumn{1}{|c|}{\underline{47.3523}}             & \multicolumn{1}{|c|}{47.1874}                             & \multicolumn{1}{|c|}{45.9614}                             & \multicolumn{1}{|c|}{\textbf{47.5429}} \\
    \multicolumn{1}{|c|}{256 views (DS = 3)} & \multicolumn{1}{|c|}{37.5167}                         & \multicolumn{1}{|c|}{41.7555}                             & \multicolumn{1}{|c|}{39.9657}
                                    & \multicolumn{1}{|c|}{44.5542}                         & \multicolumn{1}{|c|}{\underline{44.5752}}                 & \multicolumn{1}{|c|}{44.0486}                             & \multicolumn{1}{|c|}{\textbf{45.2628}} \\
    \multicolumn{1}{|c|}{192 views (DS = 4)} & \multicolumn{1}{|c|}{33.9806}                         & \multicolumn{1}{|c|}{38.5720}                             & \multicolumn{1}{|c|}{37.4422} 
                                    & \multicolumn{1}{|c|}{43.1441}                         & \multicolumn{1}{|c|}{\underline{43.1799}}                 & \multicolumn{1}{|c|}{43.0262}                             & \multicolumn{1}{|c|}{\textbf{44.0869}} \\
    \multicolumn{1}{|c|}{128 views (DS = 6)} & \multicolumn{1}{|c|}{29.9290}                         & \multicolumn{1}{|c|}{34.5357}                             & \multicolumn{1}{|c|}{34.0616} 
                                    & \multicolumn{1}{|c|}{41.2948}                         & \multicolumn{1}{|c|}{41.3474}                             & \multicolumn{1}{|c|}{\underline{41.4208}}                 & \multicolumn{1}{|c|}{\textbf{42.3326}} \\
    \multicolumn{1}{|c|}{~~96 views (DS = 8)}  & \multicolumn{1}{|c|}{27.5043}                         & \multicolumn{1}{|c|}{32.0524}                             & \multicolumn{1}{|c|}{32.0263} 
                                    & \multicolumn{1}{|c|}{39.9220}                         & \multicolumn{1}{|c|}{39.9795}                             & \multicolumn{1}{|c|}{\underline{40.1207}}                 & \multicolumn{1}{|c|}{\textbf{40.9531}} \\
    \hline \hline

    \multicolumn{1}{|c|}{\multirow{2}{*}{\small{(b) SSIM $\uparrow$}}}
                                    & \multicolumn{1}{|c|}{\multirow{2}{*}{(i) FBP}}        & \multicolumn{1}{|c|}{\multirow{1}{*}{(ii) Interp.}}       & \multicolumn{1}{|c|}{\multirow{1}{*}{(iii) MBIR}}  
                                    & \multicolumn{1}{|c|}{\multirow{1}{*}{(iv) I-Net}}      & \multicolumn{1}{|c|}{\multirow{1}{*}{(v) II-Net}}         & \multicolumn{1}{|c|}{\multirow{1}{*}{(vi) PI-Net}}   & \multicolumn{1}{|c|}{\multirow{1}{*}{(vii) Ours}}\\
                                    &                                                       & \multicolumn{1}{|c|}{\small( w$\backslash$ Cubic )}                  & \multicolumn{1}{|c|}{\small( w$\backslash$ TV )}                    
                                    & \multicolumn{1}{|c|}{\small( Level 1 )}               & \multicolumn{1}{|c|}{\small( Level 1 )}                   & \multicolumn{1}{|c|}{\small( Level 1 )}                   & \multicolumn{1}{|c|}{\small( Level 5 )} \\ \hline  
    \multicolumn{1}{|c|}{384 views (DS = 2)} & \multicolumn{1}{|c|}{0.9866}                          & \multicolumn{1}{|c|}{\underline{0.9957}}                  & \multicolumn{1}{|c|}{0.9925} 
                                    & \multicolumn{1}{|c|}{\textbf{0.9959}}                 & \multicolumn{1}{|c|}{0.9955}                              & \multicolumn{1}{|c|}{0.9947}                              & \multicolumn{1}{|c|}{0.9956} \\
    \multicolumn{1}{|c|}{256 views (DS = 3)} & \multicolumn{1}{|c|}{0.9442}                          & \multicolumn{1}{|c|}{0.9880}                              & \multicolumn{1}{|c|}{0.9764} 
                                    & \multicolumn{1}{|c|}{\underline{0.9925}}              & \multicolumn{1}{|c|}{0.9924}                              & \multicolumn{1}{|c|}{0.9922}                              & \multicolumn{1}{|c|}{\textbf{0.9928}} \\
    \multicolumn{1}{|c|}{192 views (DS = 4)} & \multicolumn{1}{|c|}{0.8884}                          & \multicolumn{1}{|c|}{0.9775}                              & \multicolumn{1}{|c|}{0.9605} 
                                    & \multicolumn{1}{|c|}{0.9899}                          & \multicolumn{1}{|c|}{0.9900}                              & \multicolumn{1}{|c|}{\underline{0.9904}}                  & \multicolumn{1}{|c|}{\textbf{0.9909}} \\
    \multicolumn{1}{|c|}{128 views (DS = 6)} & \multicolumn{1}{|c|}{0.7873}                          & \multicolumn{1}{|c|}{0.9501}                              & \multicolumn{1}{|c|}{0.9295} 
                                    & \multicolumn{1}{|c|}{0.9855}                          & \multicolumn{1}{|c|}{0.9857}                              & \multicolumn{1}{|c|}{\underline{0.9868}}                  & \multicolumn{1}{|c|}{\textbf{0.9873}} \\
    \multicolumn{1}{|c|}{~~96 views (DS = 8)}  & \multicolumn{1}{|c|}{0.7113}                          & \multicolumn{1}{|c|}{0.9196}                              & \multicolumn{1}{|c|}{0.9005} 
                                    & \multicolumn{1}{|c|}{0.9815}                          & \multicolumn{1}{|c|}{0.9818}                              & \multicolumn{1}{|c|}{\underline{0.9831}}                  & \multicolumn{1}{|c|}{\textbf{0.9837}} \\    
    \hline
    
    \end{tabular}
    \end{adjustbox}
    \caption{Quantitative comparison with respect to various numbers of views. Best and second-highest scores are in \textbf{bold} and \underline{underline}, respectively. }
    
\label{tbl:result_main}
\end{table*}

\subsection{Impact of Hierarchical Decomposition Level}
\label{sec:6.1}
In Section \ref{sec:3}, two theories regarding network performance were addressed, including the DCF theroy \cite{ye2018deep} and hierarchical measurement decomposition \cite{basus2000n2log2n}. The DCF theory \cite{ye2018deep} reveals that the working process of the neural network involves solving the regression problem with a low-rank property, and the rank is closely related to the non-zero components of the signal in the Fourier domain. To emphasize the low rankness of the signal, the study found a strong mathematical reason for the bowtie support of the measurement in the Fourier domain and utilized the hierarchical measurement decomposition \cite{basus2000n2log2n} to reduce the area of the bowtie support.

\begin{figure}[t!]
    \centering
    \includegraphics[width=1.0\textwidth]{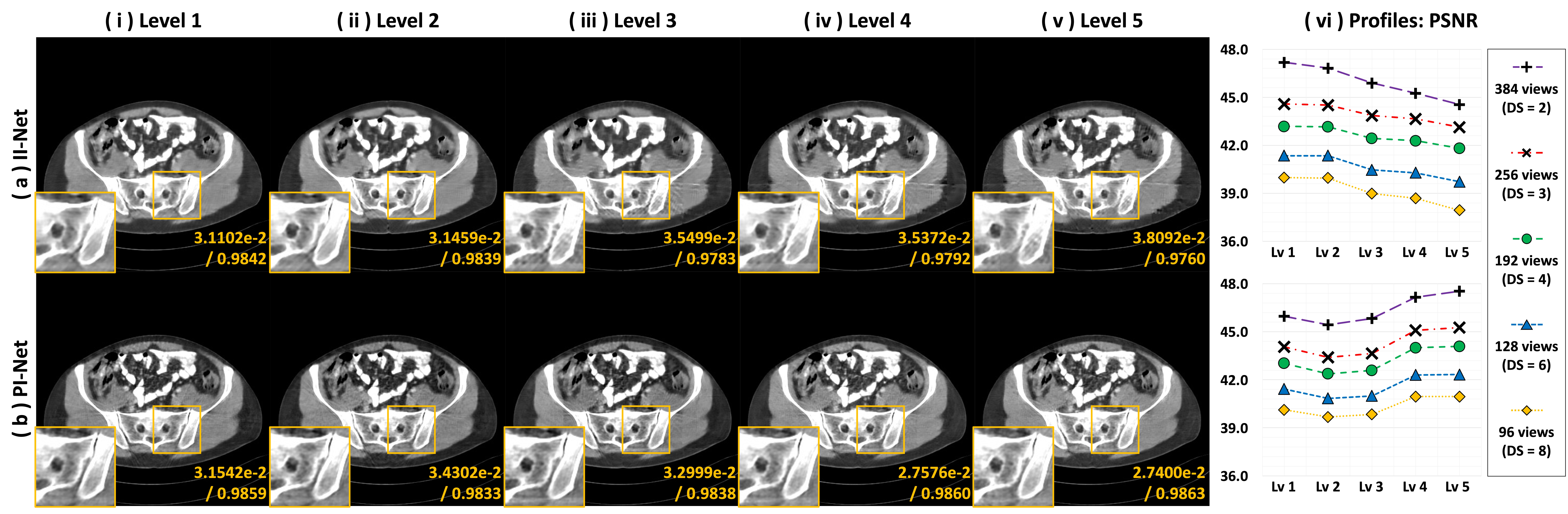}
    \caption{Reconstructed images from (a) II-Net and (b) PI-Net methods for (i-v) various decomposition levels $K$ from 1 to  5 when 128 views (DS factor $\mathcal{S} = 6$). (vi) is the PSNR profiles for various decomposition levels and views. The intensity range was set to (-160, 240) [HU]. NRMSE/SSIM values are written in the corners.}
    \label{fig:result_lvs_psnr}
\end{figure}

\begin{table*}[t!]
    \centering
    \begin{adjustbox}{width=1.0\textwidth}
    \begin{tabular}{ |c|c|c|c|c|c|c|c|c|c|c| } 
    \hline

    \multicolumn{1}{|c|}{\multirow{1}{*}{\small{(a) NRMSE$\downarrow$}}} 
                                    & \multicolumn{5}{|c|}{(i) II-Net}                                                                                                                                          & \multicolumn{5}{|c|}{(ii) PI-Net} \\ \cline{2-11}
    \multicolumn{1}{|c|}{( $10^{-2}$ )} & \multicolumn{1}{|c|}{Level 1}                 & \multicolumn{1}{|c|}{Level 2}     & \multicolumn{1}{|c|}{Level 3}     & \multicolumn{1}{|c|}{Level 4}     & \multicolumn{1}{|c|}{Level 5}     & \multicolumn{1}{|c|}{Level 1}     & \multicolumn{1}{|c|}{Level 2}     & \multicolumn{1}{|c|}{Level 3}     & \multicolumn{1}{|c|}{Level 4}                 & \multicolumn{1}{|c|}{Level 5} \\ \hline  
    \multicolumn{1}{|c|}{384 views (DS = 2)} & \multicolumn{1}{|c|}{\underline{1.5135}}  & \multicolumn{1}{|c|}{1.5791}  & \multicolumn{1}{|c|}{1.7569}  & \multicolumn{1}{|c|}{1.8911}  & \multicolumn{1}{|c|}{2.0512}  & \multicolumn{1}{|c|}{1.7470}  & \multicolumn{1}{|c|}{1.8581}  & \multicolumn{1}{|c|}{1.7725}  & \multicolumn{1}{|c|}{1.5184}              & \multicolumn{1}{|c|}{\textbf{1.4516}} \\
    \multicolumn{1}{|c|}{256 views (DS = 3)} & \multicolumn{1}{|c|}{2.0451}              & \multicolumn{1}{|c|}{2.0597}  & \multicolumn{1}{|c|}{2.2230}  & \multicolumn{1}{|c|}{2.2754}  & \multicolumn{1}{|c|}{2.4167}  & \multicolumn{1}{|c|}{2.1792}  & \multicolumn{1}{|c|}{2.3473}  & \multicolumn{1}{|c|}{2.2818}  & \multicolumn{1}{|c|}{\underline{1.9257}}  & \multicolumn{1}{|c|}{\textbf{1.8877}} \\
    \multicolumn{1}{|c|}{192 views (DS = 4)} & \multicolumn{1}{|c|}{2.4022}              & \multicolumn{1}{|c|}{2.4075}  & \multicolumn{1}{|c|}{2.6203}  & \multicolumn{1}{|c|}{2.6635}  & \multicolumn{1}{|c|}{2.8140}  & \multicolumn{1}{|c|}{2.4515}  & \multicolumn{1}{|c|}{2.6461}  & \multicolumn{1}{|c|}{2.5770}  & \multicolumn{1}{|c|}{\underline{2.1856}}  & \multicolumn{1}{|c|}{\textbf{2.1624}} \\
    \multicolumn{1}{|c|}{128 views (DS = 6)} & \multicolumn{1}{|c|}{2.9702}              & \multicolumn{1}{|c|}{2.9710}  & \multicolumn{1}{|c|}{3.2922}  & \multicolumn{1}{|c|}{3.3614}  & \multicolumn{1}{|c|}{3.5837}  & \multicolumn{1}{|c|}{2.9480}  & \multicolumn{1}{|c|}{3.1557}  & \multicolumn{1}{|c|}{3.0961}  & \multicolumn{1}{|c|}{\underline{2.6611}}  & \multicolumn{1}{|c|}{\textbf{2.6504}} \\
    \multicolumn{1}{|c|}{~~96 views (DS = 8)}  & \multicolumn{1}{|c|}{3.4822}              & \multicolumn{1}{|c|}{3.4927}  & \multicolumn{1}{|c|}{3.9075}  & \multicolumn{1}{|c|}{4.0385}  & \multicolumn{1}{|c|}{4.4029}  & \multicolumn{1}{|c|}{3.4250}  & \multicolumn{1}{|c|}{3.6107}  & \multicolumn{1}{|c|}{3.5396}  & \multicolumn{1}{|c|}{\underline{3.1137}}  & \multicolumn{1}{|c|}{\textbf{3.1117}} \\
    \hline \hline
    
    \multicolumn{1}{|c|}{\multirow{2}{*}{\small{(b) SSIM$\uparrow$}}} 
                                    & \multicolumn{5}{|c|}{(i) II-Net}                                                                                                                                          & \multicolumn{5}{|c|}{(ii) PI-Net} \\ \cline{2-11}
                                    & \multicolumn{1}{|c|}{Level 1}                 & \multicolumn{1}{|c|}{Level 2}     & \multicolumn{1}{|c|}{Level 3}     & \multicolumn{1}{|c|}{Level 4}     & \multicolumn{1}{|c|}{Level 5}     & \multicolumn{1}{|c|}{Level 1}     & \multicolumn{1}{|c|}{Level 2}     & \multicolumn{1}{|c|}{Level 3}     & \multicolumn{1}{|c|}{Level 4}                 & \multicolumn{1}{|c|}{Level 5} \\ \hline  
    \multicolumn{1}{|c|}{384 views (DS = 2)} & \multicolumn{1}{|c|}{\underline{0.9955}}  & \multicolumn{1}{|c|}{0.9952}  & \multicolumn{1}{|c|}{0.9942}  & \multicolumn{1}{|c|}{0.9936}  & \multicolumn{1}{|c|}{0.9930}  & \multicolumn{1}{|c|}{0.9947}  & \multicolumn{1}{|c|}{0.9938}  & \multicolumn{1}{|c|}{0.9945}  & \multicolumn{1}{|c|}{0.9952}              & \multicolumn{1}{|c|}{\textbf{0.9956}} \\
    \multicolumn{1}{|c|}{256 views (DS = 3)} & \multicolumn{1}{|c|}{0.9924}              & \multicolumn{1}{|c|}{0.9923}  & \multicolumn{1}{|c|}{0.9911}  & \multicolumn{1}{|c|}{0.9909}  & \multicolumn{1}{|c|}{0.9902}  & \multicolumn{1}{|c|}{0.9922}  & \multicolumn{1}{|c|}{0.9903}  & \multicolumn{1}{|c|}{0.9915}  & \multicolumn{1}{|c|}{\underline{0.9925}}  & \multicolumn{1}{|c|}{\textbf{0.9928}} \\
    \multicolumn{1}{|c|}{192 views (DS = 4)} & \multicolumn{1}{|c|}{0.9900}              & \multicolumn{1}{|c|}{0.9898}  & \multicolumn{1}{|c|}{0.9881}  & \multicolumn{1}{|c|}{0.9881}  & \multicolumn{1}{|c|}{0.9868}  & \multicolumn{1}{|c|}{0.9904}  & \multicolumn{1}{|c|}{0.9880}  & \multicolumn{1}{|c|}{0.9893}  & \multicolumn{1}{|c|}{\underline{0.9907}}  & \multicolumn{1}{|c|}{\textbf{0.9909}} \\
    \multicolumn{1}{|c|}{128 views (DS = 6)} & \multicolumn{1}{|c|}{0.9857}              & \multicolumn{1}{|c|}{0.9857}  & \multicolumn{1}{|c|}{0.9823}  & \multicolumn{1}{|c|}{0.9821}  & \multicolumn{1}{|c|}{0.9794}  & \multicolumn{1}{|c|}{0.9868}  & \multicolumn{1}{|c|}{0.9838}  & \multicolumn{1}{|c|}{0.9853}  & \multicolumn{1}{|c|}{\underline{0.9871}}  & \multicolumn{1}{|c|}{\textbf{0.9873}} \\
    \multicolumn{1}{|c|}{~~96 views (DS = 8)}  & \multicolumn{1}{|c|}{0.9818}              & \multicolumn{1}{|c|}{0.9817}  & \multicolumn{1}{|c|}{0.9768}  & \multicolumn{1}{|c|}{0.9755}  & \multicolumn{1}{|c|}{0.9700}  & \multicolumn{1}{|c|}{0.9831}  & \multicolumn{1}{|c|}{0.9801}  & \multicolumn{1}{|c|}{0.9819}  & \multicolumn{1}{|c|}{\textbf{0.9837}}     & \multicolumn{1}{|c|}{\textbf{0.9837}} \\
    \hline    
    \end{tabular}
    \end{adjustbox}
    \caption{Quantitative comparison with respect to various decomposition levels. Best and second-highest scores are in \textbf{bold} and \underline{underline}, respectively.}
\label{tbl:result_lvs}
\end{table*}

To validate the relationship between the DCF theory \cite{ye2018deep} and hierarchical measurement decomposition \cite{basus2000n2log2n}, II-Net and PI-Net were trained with datasets at various decomposition levels $K$, as shown in Figure \ref{fig:decomposition}, and the reconstructed images along with the quantitative metrics are presented in Figure \ref{fig:result_lvs_psnr} and Table \ref{tbl:result_lvs}. In particular, the PSNR profiles with respect to the decomposition levels and the number of views are depicted in Figure \ref{fig:result_lvs_psnr}(vi). Interestingly, the II-Net shows a performance degradation as the decomposition level increases, as illustrated in Figure \ref{fig:result_lvs_psnr}(a)(vi). The reason is that training the image-domain DL with entire CT images (decomposition level $K = 1$) is more advantageous than training it using image-patches (decomposition level $K > 1$) because streaking artifacts in the image-domain are defined as global artifacts rather than local artifacts. Therefore, Figure \ref{fig:result_lvs_psnr}(a) shows that it is difficult for II-Net trained with highly decomposed image-patches to clearly remove global streaking artifacts. An additional challenge in removing streaking artifacts is that 
the DL-based model is simultaneously trained on synthesized CT images with different DS ratios to ensure the generalization effect. However, performance degradation may occur if the extended data distribution does not achieve low-rank properties.
At a high decomposition level $K$, the image-domain DL is not well-trained due to various types of fragmented global streaking artifacts. The quantitative metrics are shown in Table \ref{tbl:result_lvs}(i).

\begin{figure}[t!]
    \centering
    \includegraphics[width=1.0\textwidth]{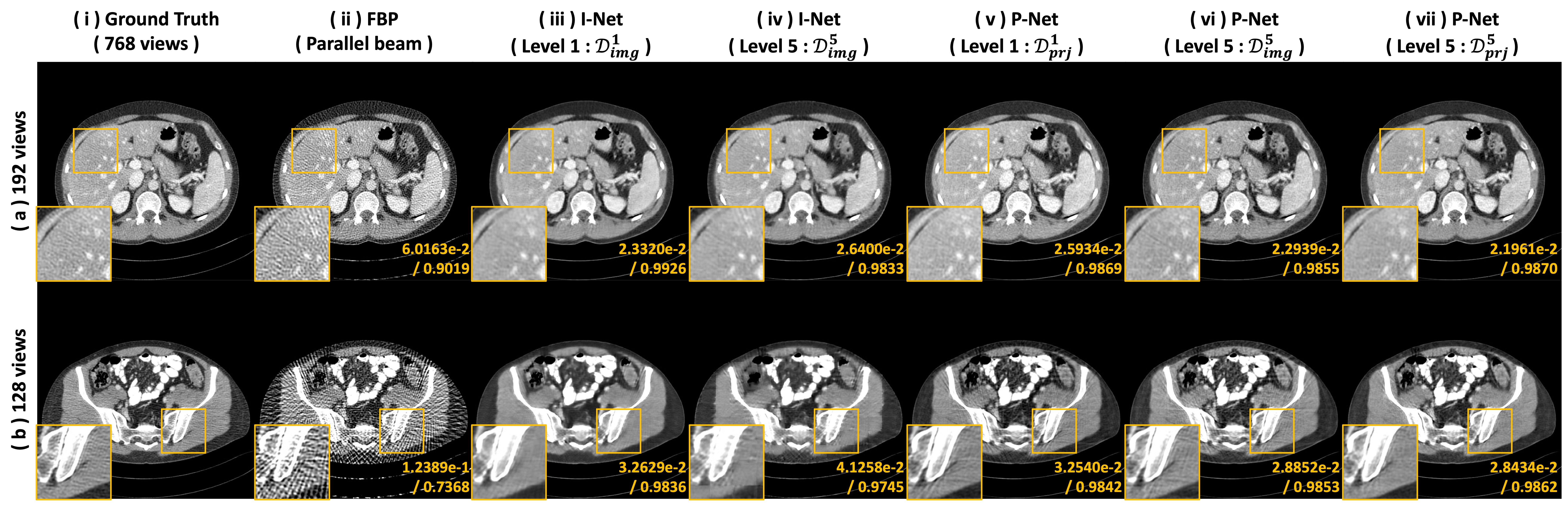}
    \caption{(i) Ground truth, (ii) FBP, and reconstructed images from (iii, iv) I-Net and (v-vii) P-Net when (a) 192 views (DS factor $S=4$) and (b) 128 views (DS factor $S=6$). (iii, iv) are the first phase networks for II-Net with the regular decomposition level $K = 1$ and $K = 5$, respectively. (v, vii) are the first phase networks for PI-Net with the hierarchical decomposition level $K = 1$ and $K = 5$, respectively. (vi) is the P-Net trained with regular decomposition method \cite{lee2018deep}. The intensity range was set to (-160, 240) [HU]. NRMSE/SSIM values are written in the corners.}
    \label{fig:result_ip_1}
\end{figure}

\begin{table}[t!]
    \centering
    \begin{adjustbox}{width=1.0\textwidth}
    \begin{tabular}{ |c|c|c|c|c|c| } 
    \hline
    PSNR $\uparrow$                          & \multicolumn{2}{|c|}{(a) I-Net}                                      & \multicolumn{3}{|c|}{(b) P-Net}                                                                                \\ \cline{2-6}
    [ dB ]                   & \multicolumn{1}{|c|}{(i) Level 1 : $\mathcal{D}_{img}^1$}& \multicolumn{1}{|c|}{(ii) Level 5 : $\mathcal{D}_{img}^5$}& \multicolumn{1}{|c|}{(i) Level 1 : $\mathcal{D}_{prj}^1$} & \multicolumn{1}{|c|}{(ii) Level 5 : $\mathcal{D}_{img}^5$}& \multicolumn{1}{|c|}{(iii) Level 5 : $\mathcal{D}_{prj}^5$} \\ \hline  
    \multicolumn{1}{|c|}{384 views (DS = 2)} & \multicolumn{1}{|c|}{47.3523}    & \multicolumn{1}{|c|}{{45.1148}}   & \multicolumn{1}{|c|}{47.0428}     & \multicolumn{1}{|c|}{47.6792}     & \multicolumn{1}{|c|}{\textbf{48.0191}} \\
    \multicolumn{1}{|c|}{256 views (DS = 3)} & \multicolumn{1}{|c|}{44.5542}    & \multicolumn{1}{|c|}{{43.1895}}   & \multicolumn{1}{|c|}{44.1547}     &\multicolumn{1}{|c|}{ 44.6508}     & \multicolumn{1}{|c|}{\textbf{45.3545}} \\
    \multicolumn{1}{|c|}{192 views (DS = 4)} & \multicolumn{1}{|c|}{43.1441}    & \multicolumn{1}{|c|}{{41.6619}}   & \multicolumn{1}{|c|}{42.8281}     & \multicolumn{1}{|c|}{43.2711}     & \multicolumn{1}{|c|}{\textbf{44.0121}} \\
    \multicolumn{1}{|c|}{128 views (DS = 6)} & \multicolumn{1}{|c|}{41.2948}    & \multicolumn{1}{|c|}{{39.4838}}   & \multicolumn{1}{|c|}{40.8534}     & \multicolumn{1}{|c|}{41.1559}     & \multicolumn{1}{|c|}{\textbf{42.1353}} \\
    \multicolumn{1}{|c|}{~~96 views (DS = 8)}& \multicolumn{1}{|c|}{39.9220}     & \multicolumn{1}{|c|}{{37.7030}}    & \multicolumn{1}{|c|}{39.4429}     & \multicolumn{1}{|c|}{39.6720}     & \multicolumn{1}{|c|}{\textbf{40.7308}} \\
    \hline 

    \end{tabular}
    \end{adjustbox}
    
    \caption{Quantitative comparison with respect to (i) I-Net and (ii) P-Net for various numbers of views and decomposition methods. Best score is in \textbf{bold}.}
\label{tbl:result_ip_metrics}
\end{table}

On the other hand, PI-Net shows the performance improvement as the decomposition level $K$ increases, as shown in Figure \ref{fig:result_lvs_psnr}(b)(vi), except for $K=1$. Table \ref{tbl:result_lvs}(ii) shows that PI-Net trained with full-size measurements (decomposition level $K = 1$) outperforms the other PI-Nets trained with measurements decomposed by the levels $K=2$ and $K=3$. However, the performance gradually improves as the decomposition level $K$ increases from the level 2 to the level 5, as shown in Figure \ref{fig:result_lvs_psnr}(b)(vi). PI-Net trained with decomposition levels $K=5$ is superior to the conventional PI-Net trained with decomposition level $K=1$. Figure \ref{fig:result_lvs_psnr}(b)(i-v) show reconstructed images from PI-Nets trained with different decomposition levels, with higher decomposition levels improving quantitative metrics, as described in Table \ref{tbl:result_lvs}(ii). Therefore, through this experiment, it was confirmed that mathematical intuition related to DCF theroy \cite{ye2018deep} and hierarchical measurement decomposition \cite{basus2000n2log2n}  was clearly and empirically verified.

\begin{figure}[t!]
    \centering
    \includegraphics[width=1.0\textwidth]{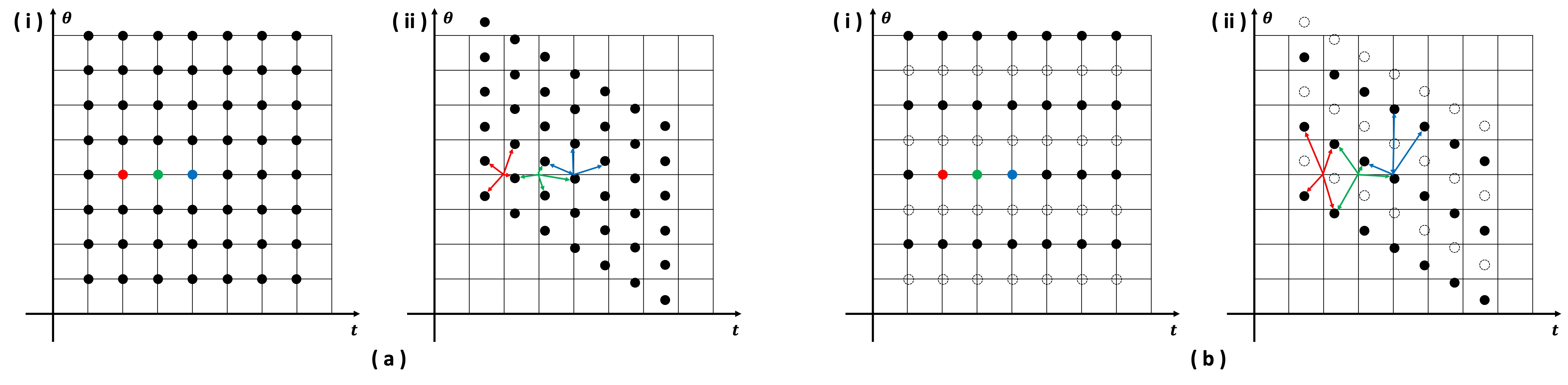}
    \caption{Measurement sampling patterns according to (i) parallel beam CT and (ii) fan beam CT. (a) and (b) are full-view and sparse-view measurements, respectively. R/G/B samples in (i) parallel beam CT are interpolated by adjacent samples indicated by R/G/B arrows in (ii) fan beam CT.}
    \label{fig:fan2para}
\end{figure}

\subsection{Performance of Data Domain}
\label{sec:6.2}

Section \ref{sec:6.1} explained why II-Net with decomposition level $K=1$ and PI-Net with decomposition level $K=5$ achieve the best performance among various decomposition levels. However, since both networks consist of two-times unrolled networks, it is difficult to determine performance differences depending on whether the data domain of the network is the image-domain or the projection-domain.
To verify the effectiveness of the data domain, I-Net (the 1st phase network of II-Net) and P-Net (the 1st phase network of PI-Net) are compared. The reconstructed images from the I-Net and the P-Net are shown in Figure \ref{fig:result_ip_1}, and the quantitative metrics are summarized in Table \ref{tbl:result_ip_metrics}.

Compared to the results of I-Nets, I-Net with level $K=1$ in Figure \ref{fig:result_ip_1}(iii) shows better removal performance of global streaking artifacts than I-Net with regular decomposition level $K=5$ in Figure \ref{fig:result_ip_1}(iv). However, P-Nets show opposite results to I-Nets. P-Net with hierarchical decomposition level $K=5$ in Figure \ref{fig:result_ip_1}(vii) outperforms P-Net with level $K=1$ in Figure \ref{fig:result_ip_1}(v) due to the narrow bowtie support in the Fourier domain. Additionally, to verify the image quality according to the decomposition method, a new P-Net was trained with level $K=5$ of regular projection patches, as shown in Figure \ref{fig:result_ip_1}(vi). The interesting point is that P-Net at level $K=5$ of regular decomposition (see Figure \ref{fig:result_ip_1}(vi)) shows similar results to P-Net at level $K=1$ (see Figure \ref{fig:result_ip_1}(v)). Through this, it can be inferred that the two networks have similar basis satisfying the low-rank properties. Considering the shift-invariant characteristic of convolution layers, the inference is reasonable. 

\subsection{Performance of Fan beam CT Geometry}
\label{sec:6.3}
\begin{figure}[t!]
    \centering
    \includegraphics[width=1.0\textwidth]{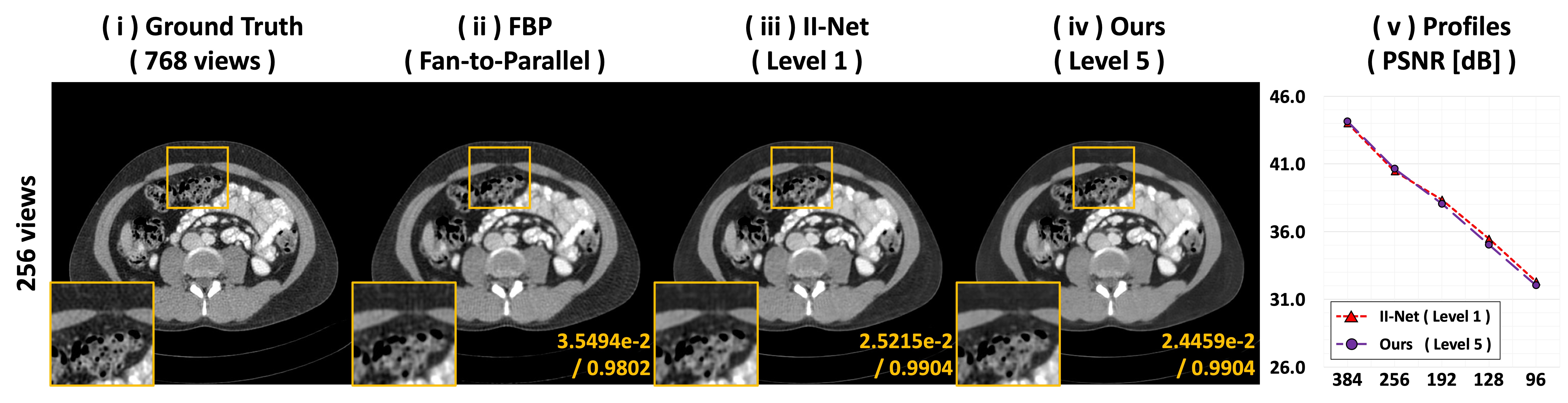}
    \caption{(i) Ground truth and (ii) FBP from fan beam CT geometry. (iii) and (iv) shows the results reconstructed from II-Net and proposed method from rebinned measurement when 256 views (DS factor $S=3$). (v) is the PSNR profiles for various number of views. The intensity range was set to (-160, 240) [HU]. NRMSE/SSIM values are written in the corners.}
    \label{fig:result_fan2para}
\end{figure}

As shown in Figure \ref{fig:decomposition} and Algorithm \ref{alg:decomposition}, the hierarchical decomposition method is defined in the parallel beam CT geometry, whereas conventional CT geometries follow equi-angular or equi-spaced CT geometries, including fan beam and cone beam CT systems. A simple way to apply the decomposition method to fan beam geometry is to transform the parallel beam projection from the fan beam CT measurements using the fan-to-parallel beam rebinning operation. Figure \ref{fig:fan2para} shows measurement sampling patterns according to (i) parallel beam and (ii) fan beam CTs. Specifically, the larger the number of views, the smaller the interpolation error in Figure \ref{fig:fan2para}(a). In other words, as shown in Figure \ref{fig:fan2para}(b), the smaller the number of views, the greater the interpolation error. The results of the fan beam CT is shown in Figure \ref{fig:result_fan2para}. The performance of our projection-domain DL is slightly degraded because the measurements were directly contaminated by interpolation errors caused by the rebinning process. However, the performance degradation can be easily compensated by using parallel beam measurements and rebinned measurements together during the training phase.

\begin{figure}[t!]
    \centering
    \includegraphics[width=1.0\textwidth]{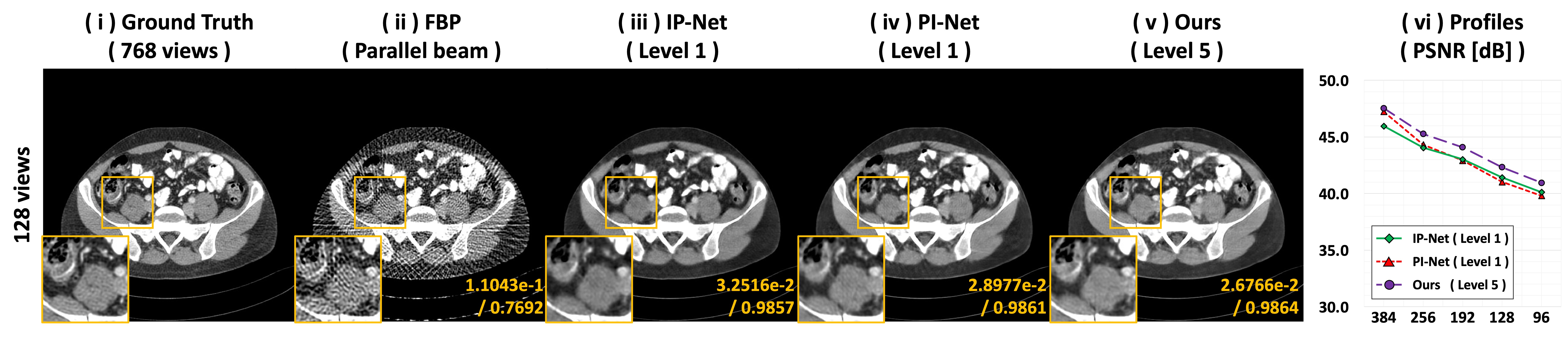}
    \caption{(i) Ground truth and (ii) FBP, and reconstructed images from (iii) IP-Net, (iv) PI-Net, and (v) proposed method with 128 views (DS factor $S = 6$). (vi) is the PSNR profiles for various number of views. The intensity range was set to (-160, 240)[HU]. NRMSE/SSIM values are written in the corners.}
    \label{fig:result_ip_pi}
\end{figure}

\subsection{Relationship between Basis Sets and Unrolled Networks}
\label{sec:6.4}
Similar to PI-Net, IP-Net, which I-Net and P-Net are sequentially connected, can be another candidate to address different data domains. 
Unfortunately, IP-Net cannot apply hierarchical decomposition method because decomposed projection data cannot be generated by applying projection operators (or Radon transforms) to image patches. Therefore, IP-Net is only trained only on datasets without decomposition. Figure \ref{fig:result_ip_pi} shows the the reconstructed results from IP-Net with level $K=1$ and PI-Net with level $K=1$ and $K=5$. As the PSNR profiles in Figure \ref{fig:result_ip_pi}(vi) shows, the PSNR trend of IP-Net with level $K=1$ is similar to that of PI-Net with level $K=1$. The experiments show empirically that both networks have similar basis sets for reconstructing CT images.

\begin{figure}[t!]
    \centering
    \includegraphics[width=1.0\textwidth]{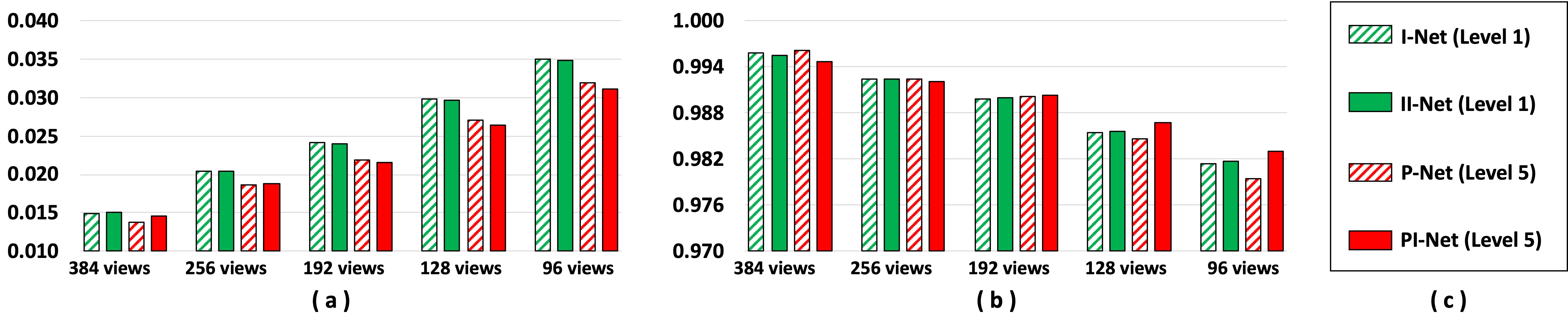}
    \caption{(a) NRMSE and (b) SSIM profiles for various number of views. (c) Identifiers used in (a) and (b). }
    \label{fig:result_lvs_quan}
\end{figure}

In this paper, two-times unrolled networks such as II-Net and PI-Net are usually used when performing experiments. 
Figure \ref{fig:result_lvs_quan} shows the NRMSE and SSIM values for sequentially connected networks. Comparing I-Net and II-Net, as shown in the green profiles in Figure \ref{fig:result_lvs_quan}, there is not much difference. However, there is a reasonable performance gap between P-Net and PI-Net as the number of views decreases. From the perspective of the network's basis based on DCF theory, I-Net and II-Net may have similar basis sets for solving regression problems using low-rank property due to the same training datasets and the same network architecture (see Figure \ref{fig:methods}(a)). Therefore, there is no meaningful performance improvement between two I-Nets. However, since P-Net and PI-Net are trained with different data domains (see Figure \ref{fig:methods}(b)), the basis sets for each network may be different. Therefore, the PI-Net can improve performance over the P-Net, a first phase network.

\section{Conclusion}


The study proposed a novel hierarchical decomposed dual-domain DL for sparse-view CT reconstruction. Conventional image-domain DL functions as an image artifact remover. However, this study reveals that there networks fail to address the underlying cause of artifacts, which is the presence of incomplete measurements. Therefore, the study uses a hierarchical decomposed projection-domain DL as a missing data reconstructor to directly confront the problems arising from incomplete measurements. For the reconstruction of undersampled projection data, the study proposed a novel projection-domain DL, which is trained with hierarchical decomposed measurements. The decomposed measurements exhibit a narrow bowtie support in the Fourier domain, thereby satisfying the low-rank property demonstrated by DCF theory. By achieving this property, the proposed method was able to outperform various conventional methods and various DLs. Furthermore, our findings revealed a direct correlation between the decomposition level and performance: the higher the decomposition level, the better the performance. 

Although the proposed method presents good theoretical justification and the experimental results, it also has some drawbacks. First, there is overestimation issue in the case of dual-domain networks including the proposed method. In weak sparse-view CT situations, the first network outperforms the second network. However, the performance of the second network is already sufficient for radiologists to make a diagnosis. Another one is the CT geometry restriction of the hierarchical decomposition method. Since the decomposition method is designed for projections measured from parallel beam CT geometry, rebinning process is required to convert fan beam CT measurements to the form of parallel beam measurement. The fan-to-parallel rebinning process is a simple operation with little computational cost, but can increase interpolation error at high downsampling levels. The performance of our projection-domain DL might be slightly degraded due to the contaminated measurement by rebinning process. However, the performance degradation can be easily compensated by using parallel beam measurements and rebinned measurements together during the training phase.


\section*{Acknowledgements}
This research was supported by the MSIT(Ministry of Science and ICT), Korea, under the ITRC(Information Technology Research Center) support program(IITP-2024-2020-0-01602) supervised by the IITP(Institute for Information \& Communications Technology Planning \& Evaluation).

\clearpage

\section*{References}



\bibliographystyle{unsrt}
\bibliography{main.bib}

\end{document}